\PassOptionsToPackage{table}{xcolor}
\documentclass[sigconf, screen]{acmart}

\usepackage{multirow}
\usepackage{booktabs}
\usepackage{xcolor}
\usepackage{enumitem}
\usepackage[skins,breakable]{tcolorbox}

\definecolor{sysfill}{RGB}{232,240,254}
\definecolor{systitle}{RGB}{200,218,248}
\definecolor{userfill}{RGB}{254,241,230}
\definecolor{usertitle}{RGB}{248,225,205}
\definecolor{outfill}{RGB}{232,247,232}
\definecolor{outtitle}{RGB}{210,238,210}
\definecolor{judgefill}{RGB}{242,235,250}
\definecolor{judgetitle}{RGB}{225,215,242}
\definecolor{feedfill}{RGB}{242,242,242}
\definecolor{feedtitle}{RGB}{225,225,225}

\AtBeginDocument{%
  }

\copyrightyear{2026}
\acmYear{2026}
\setcopyright{cc}
\setcctype{by}
\acmConference[MM '26]{Proceedings of the 34th ACM International Conference on Multimedia}{November 10--14, 2026}{Rio de Janeiro, Brazil}
\acmBooktitle{Proceedings of the 34th ACM International Conference on Multimedia (MM '26), November 10--14, 2026, Rio de Janeiro, Brazil}
\acmDOI{10.1145/3767308.3835924}
\acmISBN{979-8-4007-2213-4/2026/11}

\begin{document}

\title{HalluScope: Fine-grained Hallucination Diagnosis for Multimodal Large Language Models}

\author{Weilin Jin}
\affiliation{%
  \institution{Peking University}
  \city{Beijing}
  \country{China}
}
\email{2401112012@stu.pku.edu.cn}

\author{Mingyu Wang}
\affiliation{%
  \institution{Peking University}
  \city{Beijing}
  \country{China}
}
\email{2501210702@stu.pku.edu.cn}

\author{Wenbo Li}
\authornote{Project leader.}
\affiliation{%
  \institution{Joy Future Academy}
  \city{Beijing}
  \country{China}
}
\email{fenglinglwb@gmail.com}

\author{Haoyang Huang}
\affiliation{%
  \institution{Joy Future Academy}
  \city{Beijing}
  \country{China}
}
\email{huanghaoyang@jd.com}

\author{Yifan Wu}
\authornote{Corresponding author.}
\affiliation{%
  \institution{Peking University}
  \city{Beijing}
  \country{China}
}
\email{yifanwu@pku.edu.cn}

\author{Ying Li}
\authornotemark[2]
\affiliation{%
  \institution{Peking University}
  \city{Beijing}
  \country{China}
}
\email{li.ying@pku.edu.cn}

\author{Gang Huang}
\affiliation{%
  \institution{Peking University}
  \city{Beijing}
  \country{China}
}
\email{hg@pku.edu.cn}

\author{Zhonghai Wu}
\affiliation{%
  \institution{Peking University}
  \city{Beijing}
  \country{China}
}
\email{wuzh@pku.edu.cn}

\renewcommand{\shortauthors}{Jin et al.}

\begin{abstract}

Although Multimodal Large Language Models have achieved strong performance across a wide range of vision-language tasks, they still suffer from hallucinations, where model outputs become inconsistent with the visual content, textual context, or commonsense knowledge. Existing studies primarily address this problem through coarse-grained detection. However, these approaches often provide insufficient diagnostic information for understanding hallucination types and supporting downstream hallucination mitigation. To bridge this gap, we propose fine-grained hallucination diagnosis for MLLMs, a new unified task that jointly performs hallucination detection, classification, and interpretable explanation generation. We develop an automated data generation pipeline and construct HalluScope-30K, a large-scale diagnostic dataset covering eight sources and five task categories. Based on this dataset, we design a multi-granular joint reward function and train two diagnosis models, HalluScope-4B and HalluScope-8B, which achieve state-of-the-art performance on both the MHALO benchmark and our fine-grained hallucination classification benchmark. Notably, detection and classification are mutually beneficial under joint optimization. Furthermore, diagnosis-driven feedback experiments show that the fine-grained diagnostic explanations produced by our model effectively guide target models to correct their hallucinations, with full diagnosis substantially outperforming all baselines on both Qwen3-VL-8B-Instruct and LLaVA-1.5-7B. Our code, data, and models are available at \url{https://github.com/wkinglin/HalluScope}.

\end{abstract}



\maketitle

\section{INTRODUCTION}

Multimodal Large Language Models (MLLMs) \cite{bai2025qwen3,zhu2023minigpt,glm2024chatglm,li2023blip} have achieved strong performance across a wide range of vision-language tasks, including image understanding \cite{chen2015microsoft}, visual question answering \cite{antol2015vqa,kuang2025natural}, and visual reasoning \cite{wang2024exploring,peng2023kosmos}. Despite these advances, hallucination remains a persistent challenge, as model outputs may become inconsistent with visual content, textual context, or commonsense knowledge \cite{bai2024hallucination,liu2024survey,sahoo2024comprehensive}. Such errors undermine the reliability of MLLM outputs, particularly in high-stakes scenarios such as healthcare, autonomous driving, and scientific analysis \cite{li2023llavamed,chu2025reducing}. Consequently, identifying and mitigating hallucinations has become a critical challenge for MLLMs.

Existing studies primarily address hallucination through coarse-grained detection or feedback, typically formulating it as a binary classification problem~\cite{chen2024unified,zhang2025dhcp,yu2024rlhf}. However, such methods only determine whether a response contains hallucinations, without localizing hallucinated spans, identifying their types, or explaining the underlying causes, thereby providing insufficient diagnostic information for effective hallucination analysis and mitigation.

To address these limitations, recent work has explored Fine-grained Hallucination Detection (FHD)~\cite{cai2025mhalo}, which localizes hallucinated spans within the response at the token level. Compared with coarse-grained methods, FHD provides more informative signals for hallucination analysis, enabling high-quality data annotation~\cite{yu2024rlhf,gunjal2024mhal}, model alignment~\cite{fu2024tldr}, and feedback generation.

Despite these advances, existing FHD methods still focus primarily on span-level detection, without further analyzing the types or causes of identified hallucinations. Yet different hallucination types arise from fundamentally different failure mechanisms~\cite{mishra2024fine} and require distinct mitigation strategies. For instance, \textbf{Perceptual Hallucinations} such as object recognition errors can often be mitigated by re-examining the images, whereas \textbf{Reasoning Hallucinations} involving spatial reasoning or complex semantic relations require more sophisticated verification mechanisms. Fine-grained hallucination classification can reveal such error patterns, enabling targeted analysis and more effective mitigation. Beyond classification, interpretable explanations that describe why a hallucination occurs and how it may be corrected provide actionable guidance for downstream correction. Therefore, detection and classification are closely related and can reinforce each other under joint optimization, while interpretable explanations provide additional guidance for downstream hallucination correction.

Motivated by these observations, we propose \textbf{fine-grained hallucination diagnosis} for MLLMs, a task that unifies hallucination detection, classification, and interpretable explanation generation. To support this task, we develop a multimodal hallucination diagnosis data generation pipeline and construct \textbf{HalluScope-30K}, a large-scale dataset for this task. Based on this dataset, we design a multi-granular joint reward function to jointly optimize hallucination detection and classification, and train two diagnosis models, \textbf{HalluScope-4B} and \textbf{HalluScope-8B}. 

Since hallucination diagnosis is inherently a multi-stage task, we evaluate each stage independently. For fine-grained hallucination detection, we evaluate our models on the MHALO benchmark. Compared with representative MLLMs and specialized detection baselines, HalluScope achieves state-of-the-art performance, with an average $\mathbf{F1}_{\mathrm{M}}$ of \textbf{64.03} and $\mathbf{F1}_{\mathrm{IoU}}$ of \textbf{57.57}. For hallucination classification, we construct a fine-grained benchmark spanning five task categories. HalluScope attains an $\mathbf{F1}_{\mathrm{Macro}}$ of \textbf{51.66} and an $\mathbf{F1}_{\mathrm{Micro}}$ of \textbf{59.60}, while also generalizing well to the out-of-distribution HalLoc benchmark. To evaluate hallucination mitigation, we conduct diagnosis-driven feedback experiments on two target models. The results show that finer-grained feedback consistently yields greater improvements: full diagnosis raises accuracy from \textbf{68.76\%} to \textbf{81.17\%} on Qwen3-VL-8B-Instruct and from \textbf{33.02\%} to \textbf{50.34\%} on LLaVA-1.5-7B, with its mitigation benefits further validated on the external AMBER benchmark. Overall, these results demonstrate that detection and classification reinforce each other under joint optimization, while richer diagnostic feedback enables more effective hallucination mitigation. Our contributions are as follows:

\begin{itemize}
  \item We propose a new task, \textbf{fine-grained hallucination diagnosis} for MLLMs, which unifies hallucination detection, classification, and interpretable explanation generation within a single diagnostic framework.
  \item We develop an automated data generation pipeline and construct \textbf{HalluScope-30K}, a large-scale dataset that provides fine-grained supervision for hallucination diagnosis.
  \item We design a multi-granular joint reward function that enables detection and classification to benefit each other during training, and train two diagnosis models, \textbf{HalluScope-4B} and \textbf{HalluScope-8B}.
  \item We construct a fine-grained benchmark for hallucination classification, where HalluScope achieves the best performance, and further show that finer-grained diagnostic feedback consistently improves hallucination mitigation across two target models.
\end{itemize}

\begin{figure}[t]
  \centering
  \includegraphics[width=\columnwidth]{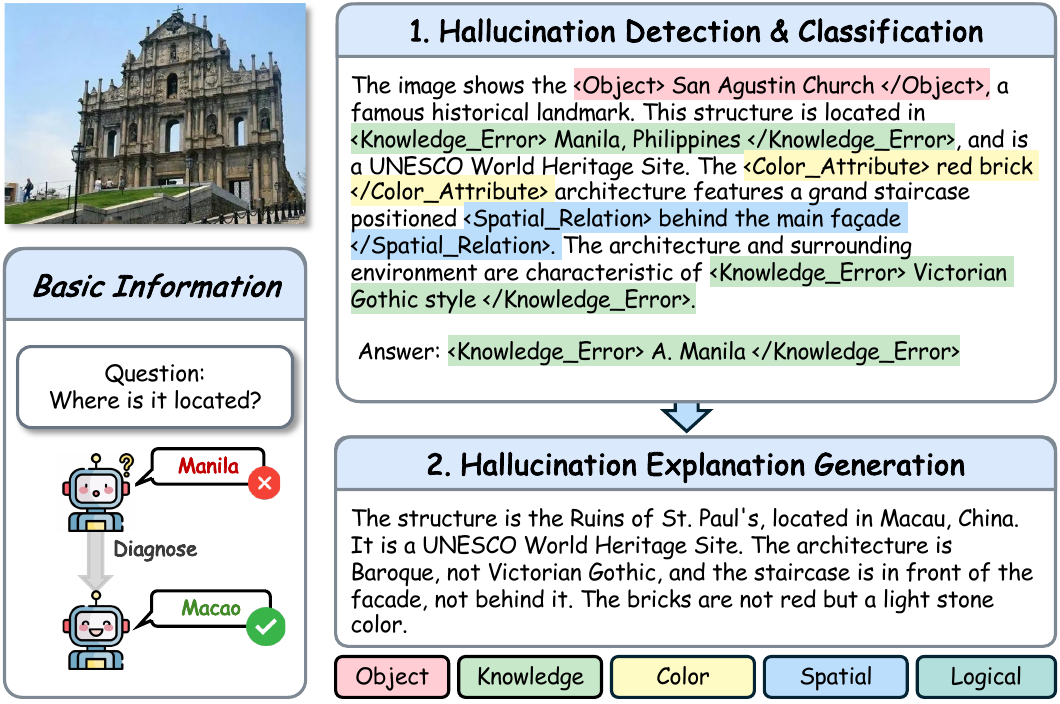}
  \caption{Example of fine-grained hallucination diagnosis. Given an image-question pair and an MLLM response, HalluScope (1) detects hallucinated spans with type classification, and (2) generates explanations with correction suggestions.}
  \label{fig:case}
  \Description{A hallucination diagnosis example showing detection, classification, and explanation generation for hallucinated spans in an MLLM response.}
\end{figure}

\section{RELATED WORK}

\subsection{Hallucination in MLLMs}

Hallucinations in multimodal large language models (MLLMs), characterized by inconsistencies between generated content and the visual input, textual context, or established world knowledge~\cite{liu2024survey,chen2026survey}, have attracted increasing attention. Such inconsistencies are commonly attributed to strong language priors and statistical biases inherited from pre-training data~\cite{leng2024mitigating}.

Prior studies address hallucination from two complementary angles: detection, which identifies inconsistencies between model outputs and visual or textual evidence~\cite{wang2023evaluation,zhai2023halle,chen2024unified,sahu2024pelican,zhang2024vl}, and mitigation, which reduces hallucination through various training-time and inference-time strategies~\cite{sun2024aligning,zhao2023beyond,yu2024rlhf,yin2024woodpecker,leng2024mitigating,zhu2025ibd,wang2024mitigating,liu2025reducing,yang2025nullu}. While effective mitigation often relies on accurate detection signals, most existing methods treat detection and mitigation as separate stages without providing unified diagnostic feedback that integrates error localization, type classification, and causal explanation. Our work addresses this gap by proposing fine-grained hallucination diagnosis as a unified task.

\subsection{Fine-grained Hallucination Detection and Mitigation for MLLMs}

Recent studies increasingly move toward fine-grained analysis for more precise hallucination supervision. Early efforts decompose model outputs into smaller textual or semantic units such as clauses, sentences, or claims~\cite{gunjal2024mhal,xiao2025fine_grained_feedback,sahu2024pelican}. More recent work pushes detection and mitigation toward token-level modeling~\cite{cai2025mhalo, whitehead2024pre, fieback2024metatoken, fu2024tldr, fieback2025ecd, zollicoffer2025mtre}, formulating hallucination detection as sequence labeling or estimating token-level hallucination likelihood through meta-classification, reward modeling, and decoding-time estimation. Among them, MHALO~\cite{cai2025mhalo} trains a dedicated model for token-level localization with coarse-grained type prediction. Nevertheless, existing methods remain primarily focused on locating hallucinated spans, while offering limited type granularity and little diagnostic explanation for downstream correction. As a result, their outputs provide limited actionable guidance for hallucination mitigation.

\section{PRELIMINARY}

\subsection{Hallucination Diagnosis Task Definition}
We formulate fine-grained hallucination diagnosis as a unified task that encompasses hallucination detection, hallucination classification, and interpretable explanation generation. Given a multimodal input consisting of an image, a textual query, and the model's response, the task aims to localize hallucinated spans in the response, classify each span into a predefined hallucination type, and generate interpretable explanations that describe the causes of hallucination and suggest possible correction strategies.

Specifically, for hallucination detection, the model identifies and marks hallucinated spans within the response through structured XML annotations. For hallucination classification, we adopt the taxonomy proposed in MHALO~\cite{cai2025mhalo}, which organizes hallucinations into two high-level categories: \textit{perception} and \textit{reasoning}. Perception-level hallucinations arise from errors in visual understanding or information extraction from images, whereas reasoning-level hallucinations originate from incorrect inference or logical reasoning based on perceived information. For explanation generation, the model produces natural language descriptions that explain why each hallucination occurs and suggest how it may be corrected.

\subsection{Hallucination Diagnosis Formulation}
\label{sec:prelim-notation}

\noindent\textbf{Notation for data construction:}
In the data generation pipeline, each sample is represented as
$x_k = (i_k, q_k, g_k, o_k, y_k^{*})$, where $i_k$ denotes the visual
input, $q_k$ the question, and $g_k$ the
ground-truth answer. Given $(i_k, q_k)$, the base model first generates a
response $o_k \sim p_{\theta}(\cdot \mid i_k, q_k)$. Based on the consistency
between $o_k$ and $g_k$, the response is further processed through
hallucination injection or hallucination annotation to produce the final
hallucination-diagnosed response $y_k^{*}$. Each diagnosed span in $y_k^{*}$ is assigned one of $C$ predefined hallucination types.

\noindent\textbf{Notation for evaluation:}
To evaluate hallucination diagnosis, we denote the set of predicted spans as
$P=\{p_i\}_{i=1}^{n}$ and the set of ground-truth spans as
$G=\{g_j\}_{j=1}^{m}$, where $n$ and $m$ denote the numbers of predicted and
ground-truth spans, respectively. Each span corresponds to a contiguous token
segment in the response and is represented by its start and end boundaries:
\[
p_i = [s_i^p, e_i^p], \qquad g_j = [s_j^g, e_j^g],
\]
where $s_i^p,e_i^p$ and $s_j^g,e_j^g$ denote the start and end positions of the $i$-th predicted span and the $j$-th ground-truth span, respectively. The span lengths are
\[
|p_i| = e_i^p - s_i^p + 1, \qquad |g_j| = e_j^g - s_j^g + 1.
\]
We further use $T(p_i)$ and $T(g_j)$ to denote the token sets contained in the predicted span $p_i$ and the ground-truth span $g_j$, respectively, and their token overlap is defined as $|T(p_i)\cap T(g_j)|$.

\section{METHOD}
\begin{figure*}[t]
  \centering
  \includegraphics[width=\textwidth]{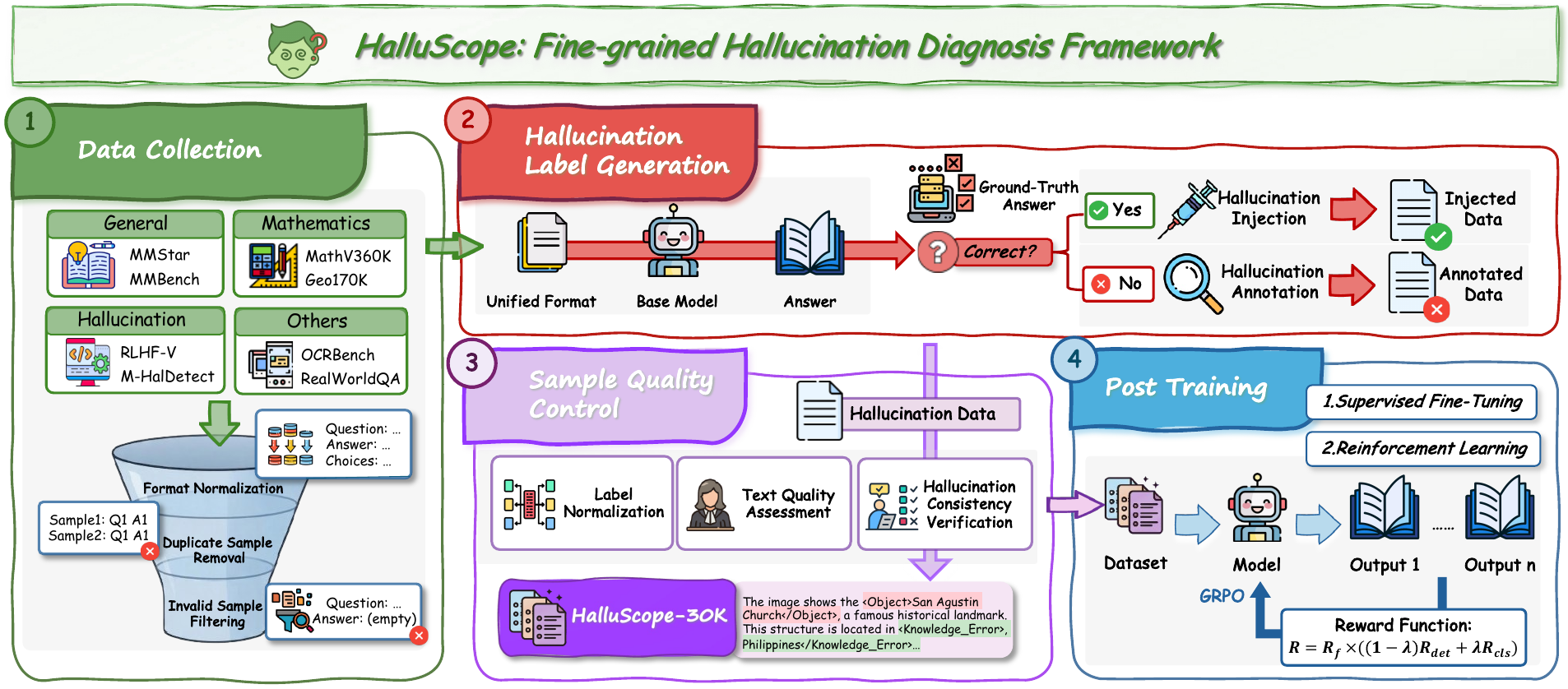}
  \caption{Overview of the HalluScope data generation and training framework.}
  \label{fig:overview}
  \Description{An overview figure of the HalluScope framework, including multi-source data collection, response generation, hallucination injection and annotation, quality control, dataset construction, and the training of HalluScope-4B and HalluScope-8B with a multi-granular joint reward function.}
\end{figure*}

Figure~\ref{fig:overview} presents the HalluScope framework. Given the collected data, the framework generates model responses and constructs fine-grained hallucination labels through complementary \textit{injection} and \textit{annotation} strategies. After multi-stage quality control, it yields HalluScope-30K. Based on this dataset, we train two diagnosis models, HalluScope-4B and HalluScope-8B.

\subsection{Pipeline of Data Generation}

\subsubsection{Data Collection} 

The HalluScope framework begins with the collection of eight datasets spanning five task categories: (1) \textit{general perception}: MMBench~\cite{liu2024mmbench} and MMStar~\cite{chen2024mmstar}; (2) \textit{mathematical reasoning}: Geo170K~\cite{gao2023g} and MathV360K~\cite{shi2024math}; (3) \textit{OCR}: OCRBench~\cite{liu2024ocrbench}; (4) \textit{spatial reasoning}: RealWorldQA~\cite{xai2024grok}; and (5) \textit{hallucination detection}: M-HalDetect~\cite{gunjal2024mhal} and RLHF-V~\cite{yu2024rlhf}. Since these datasets differ in format and annotation, we first perform standardized preprocessing: unifying data fields, removing duplicates, and filtering invalid samples (e.g., missing answers). Each processed sample follows the notation $x_k$ defined in Section~\ref{sec:prelim-notation}.

Given the unified dataset, we use the base model to generate a response for each sample. Since the original ground-truth answers are typically brief and lack sufficient detail for fine-grained annotation, we regenerate longer, more informative responses to provide richer context for hallucination diagnosis. Unlike approaches that expand ground-truth answers with strong closed-source models, constructing hallucination data from the base model's actual outputs better preserves the realism and diversity of the samples, as they reflect the model's genuine hallucination behaviors across different tasks rather than artificially simulated ones.

\subsubsection{Hallucination Label Generation} 

After obtaining model responses, we construct fine-grained hallucination labels by partitioning samples based on the consistency between the model response and the ground-truth answer. For \textbf{correctly answered samples}, we employ a closed-source model (Gemini-3-Pro) to perform \textit{hallucination injection}: analyzing the original response and introducing targeted modifications tailored to specific hallucination types while maintaining contextual coherence. The injected hallucinations are constrained to be diverse across types and natural within context, ensuring broad coverage of the hallucination taxonomy. For \textbf{incorrectly answered samples}, we employ the same closed-source model to perform \textit{hallucination annotation}: localizing hallucinated spans in the response and assigning corresponding type labels. Since these responses already contain genuine hallucinations produced by the base model, annotation directly captures realistic hallucination behaviors without artificial simulation. By combining both strategies, we preserve hallucination samples from genuine model outputs while also synthesizing new ones from correct responses, jointly yielding fine-grained supervision signals that specify both hallucination locations and types.

\subsubsection{Sample Quality Control} 

Finally, we perform quality control through a three-stage pipeline: (1) \textit{label normalization}, which maps non-standard type labels to the standardized set of 12 hallucination types through rule-based mappings (e.g., case normalization and synonym resolution); (2) \textit{text quality assessment}, which uses an LLM-as-a-judge approach to score clarity, fluency, and coherence, retaining only samples that exceed predefined thresholds; and (3) \textit{hallucination consistency verification}, which checks format correctness, type validity, and contextual consistency of the generated labels. Only samples passing all three stages are retained, yielding HalluScope-30K with nearly 30,000 samples across five task categories and eight sources, each with fine-grained hallucination diagnosis labels.

\subsection{Training of HalluScope Models}
\label{Post-Training}

Based on HalluScope-30K, we train two diagnosis models through a two-stage paradigm: supervised fine-tuning (Section~\ref{sec:sft}) followed by reinforcement learning (Section~\ref{sec:rl}).

\subsubsection{Supervised Fine-Tuning}
\label{sec:sft}

In the supervised fine-tuning stage, we use HalluScope-30K as supervision to train the base model with LoRA~\cite{hu2022lora}, enabling it to generate structured diagnostic outputs that unify hallucination detection, classification, and interpretable feedback generation. This stage serves as a cold-start initialization for subsequent reinforcement learning: the model learns to follow the required output format and acquires preliminary hallucination detection and classification abilities, reducing ineffective exploration during RL optimization~\cite{zhang2025openmmreasonerpushingfrontiersmultimodal}.

\subsubsection{Reinforcement Learning}
\label{sec:rl}

We then apply Group Relative Policy Optimization (GRPO)~\cite{guo2025deepseek} for reinforcement learning. For each sample $x_k$, the policy generates a group of $K$ candidate outputs, each scored by a multi-granular reward $R(y)$ that jointly evaluates format compliance, span-level detection quality, and type classification accuracy. Using the notation defined in Section~\ref{sec:prelim-notation}, the reward for a candidate output $y$ is formulated as:
\begin{equation}
R(y)=\mathbb{I}_{\mathrm{format}}\left((1-\lambda)\, r_{\mathrm{det}}(y)+\lambda\, r_{\mathrm{cls}}(y)\right)
\end{equation}
\noindent\textbf{Format Reward.} $\mathbb{I}_{\text{format}}$ is a binary indicator that equals 1 only when the output satisfies all formatting constraints: the output must be enclosed within \texttt{<Tagged\_Text>} tags, each hallucinated span annotated with a \texttt{<hallucination type="TYPE">} tag following XML syntax, and every type label must belong to the predefined taxonomy. This gating ensures that only structurally valid outputs receive positive diagnostic rewards.

\noindent\textbf{Hallucination Detection Reward.} The detection reward is a core component of the overall objective. Since the fine-grained hallucination diagnosis task requires the model not only to identify all hallucinations, but also to localize each hallucinated span accurately, the detection reward is designed to jointly consider both coverage and precision of the predicted spans. To this end, we define a composite detection reward as:
\begin{equation}
r_{\mathrm{det}}(y) = r_{\mathrm{content}}(y) \cdot r_{\mathrm{count}}(y)
\label{eq:detection_reward}
\end{equation}
Here $r_{\mathrm{content}}$ computes the average IoU over optimal span pairs matched by the Hungarian algorithm:
\begin{equation}
r_{\mathrm{content}}(y)=\frac{1}{|G|}\sum_{(i,j)\in\hat{M}}\mathrm{IoU}(p_i,g_j)
\label{eq:content_reward}
\end{equation}
where $\hat{M}$ maximizes the total IoU across all matched span pairs. The count reward $r_{\mathrm{count}}$ measures the consistency between the number of predicted and ground-truth spans. Since content matching alone may not penalize models that produce too many or too few spans, we introduce a Gaussian kernel that yields a maximum reward of 1 when the counts match and decays as the discrepancy increases:
\begin{equation}
r_{\mathrm{count}}(y) =
\exp\left(-\frac{(n-m)^2}{m}\right),
\label{eq:count_reward}
\end{equation}
where $n=|P|$ and $m=|G|$ denote the numbers of predicted and ground-truth hallucinated spans, respectively.

\noindent\textbf{Hallucination Classification Reward.}
Since classification builds upon successful detection, the classification reward reuses the optimal matching $\hat{M}$ and rewards matched pairs whose predicted and ground-truth types agree:
\begin{equation}
r_{\mathrm{cls}}(y) = \frac{1}{|G|} \sum_{(i,j)\in\hat{M}} \mathrm{IoU}(p_i,g_j)\cdot \mathbf{1}[t_i^p=t_j^g]
\label{eq:classification_reward}
\end{equation}
Overall, this hierarchical reward jointly optimizes output structure, span localization, and type classification, reducing reward sparsity and improving training stability compared with a single reward based only on the final matching outcome.

\section{EXPERIMENTS}


\begin{table*}[t]
\centering
\caption{Results on the MHALO benchmark. $\mathbf{F1}_{\mathrm{M}}$ and
$\mathbf{F1}_{\mathrm{IoU}}$ measure the quality of span-level
hallucination detection, while IF denotes the format validity rate.
Best results are in bold, and second-best results are underlined.}
\label{tab:results_mhalo}
\setlength{\tabcolsep}{4.6pt}
\renewcommand{\arraystretch}{1.12}
\small
\resizebox{\textwidth}{!}{
\begin{tabular}{lccccccccccccccc}
\toprule
& \multicolumn{3}{c}{\textbf{RLHF-V (500)}} 
& \multicolumn{3}{c}{\textbf{M-HalDetect (500)}} 
& \multicolumn{3}{c}{\textbf{Geo170K (500)}} 
& \multicolumn{3}{c}{\textbf{MathV360K (500)}} 
& \multicolumn{3}{c}{\textbf{Average}} \\
\cmidrule(lr){2-4}
\cmidrule(lr){5-7}
\cmidrule(lr){8-10}
\cmidrule(lr){11-13}
\cmidrule(lr){14-16}
\textbf{Model} 
& $\mathbf{F1}_{\mathrm{M}}$ & $\mathbf{F1}_{\mathrm{IoU}}$ & IF 
& $\mathbf{F1}_{\mathrm{M}}$ & $\mathbf{F1}_{\mathrm{IoU}}$ & IF 
& $\mathbf{F1}_{\mathrm{M}}$ & $\mathbf{F1}_{\mathrm{IoU}}$ & IF 
& $\mathbf{F1}_{\mathrm{M}}$ & $\mathbf{F1}_{\mathrm{IoU}}$ & IF 
& $\mathbf{F1}_{\mathrm{M}}$ & $\mathbf{F1}_{\mathrm{IoU}}$ & IF \\
\midrule
\multicolumn{16}{c}{\textbf{Baselines}} \\
\midrule
Qwen3-VL-4B-Instruct 
& 33.81 & 27.36 & 99.6
& 16.33 & 13.06 & 95.2
& 40.74 & 26.05 & 94.8
& 41.69 & 39.46 & 87.6 
& 33.14 & 26.48 & 94.3 \\
Qwen3-VL-8B-Instruct 
& 40.30 & 34.62 & 99.6
& 18.61 & 13.69 & 99.0
& 37.40 & 24.46 & 86.8 
& 51.12 & 48.93 & 89.0
& 36.86 & 30.43 & 93.6 \\
Gemini-3-Flash
& 53.20 & \underline{43.26} & 100.0
& 45.25 & 36.45 & 100.0 
& 63.95 & 48.55 & 100.0 
& 74.87 & 66.06 & 100.0 
& 59.32 & 48.58 & 100.0 \\
Gemini-3-Pro 
& 50.71 & 40.68 & 100.0 
& \textbf{48.37} & \underline{41.61} & 100.0 
& 63.14 & 46.96 & 100.0 
& 72.20 & 62.13 & 100.0 
& 58.61 & 47.84 & 100.0 \\
GPT-4o 
& 45.83 & 34.23 & 100.0 
& 36.11 & 32.28 & 100.0 
& 57.38 & 42.58 & 99.6 
& 69.49 & 59.53 & 99.8 
& 52.20 & 42.16 & 99.8 \\
GPT-5 
& 51.26 & 41.32 & 100.0
& 40.31 & 31.80 & 100.0
& 53.19 & 38.08 & 100.0
& 63.53 & 49.52 & 100.0
& 52.07 & 40.18 & 100.0 \\
HALODET-4B$^\dagger$
& 35.43 & 28.66 & 100.0
& 27.41 & 24.66 & 99.4
& 44.34 & 34.58 & 80.2
& 73.82 & 68.82 & 97.6
& 45.25 & 39.18 & 94.3 \\
HALODET-8B$^\dagger$
& 37.94 & 30.52 & 99.8
& 20.34 & 18.92 & 100.0
& 35.13 & 26.43 & 83.4
& 66.61 & 61.36 & 98.2
& 40.00 & 34.31 & 95.3 \\
\midrule
\multicolumn{16}{c}{\textbf{Our Models}} \\
\midrule

\textbf{HalluScope-4B}
& \underline{54.63} & \textbf{43.66} & 100.0
& 44.36 & 39.36 & 100.0
& \underline{68.37} & \underline{64.14} & 99.0
& \underline{77.77} & \underline{73.11} & 99.4
& \underline{61.28} & \underline{55.07} & 99.6 \\

\textbf{HalluScope-8B}
& \textbf{55.17} & \textbf{43.66} & 100.0
& \underline{47.24} & \textbf{45.31} & 99.8
& \textbf{74.65} & \textbf{67.05} & 99.4
& \textbf{79.06} & \textbf{74.26} & 99.0
& \textbf{64.03} & \textbf{57.57} & 99.6 \\
\bottomrule
\end{tabular}
}
\vspace{2pt}
{\footnotesize $^\dagger$ Since the model weights are not publicly available, we retrain HALODET following the original paper's method and data, using the same base models as our approach.}
\end{table*}


\begin{table}[t]
\centering
\small
\caption{Main results on our hallucination classification benchmark. $\mathbf{F1}_{\mathrm{M}}$ and $\mathbf{F1}_{\mathrm{IoU}}$ evaluate span-level detection, while $\mathbf{F1}_{\mathrm{Macro}}$ and $\mathbf{F1}_{\mathrm{Micro}}$ evaluate type classification. Best results are in bold, and second-best results are underlined.}
\label{tab:results_ours}
\setlength{\tabcolsep}{4.6pt}
\renewcommand{\arraystretch}{1.12}
\begin{tabular}{lcccc}
\toprule
\textbf{Model} & $\mathbf{F1}_{\mathrm{M}}$ & $\mathbf{F1}_{\mathrm{IoU}}$ & $\mathbf{F1}_{\mathrm{Macro}}$ & $\mathbf{F1}_{\mathrm{Micro}}$ \\
\midrule

Qwen3-VL-4B-Instruct & 44.03 & 36.91 & 27.55 & 35.27 \\
Qwen3-VL-8B-Instruct & 56.47 & 49.27 & 33.73 & 40.28 \\
Gemini-3-Flash & 67.33 & 60.45 & 44.44 & 47.87 \\
Gemini-3-Pro & 67.55 & 61.17 & \underline{49.14} & 53.14 \\
GPT-4o & 61.67 & 55.30 & 40.10 & 48.12 \\
GPT-5 & 62.70 & 55.21 & 43.93 & 43.54 \\

\midrule
\textbf{HalluScope-4B} & \underline{69.76} & \underline{64.77} & 48.46 & \underline{55.79} \\
\textbf{HalluScope-8B} & \textbf{72.20} & \textbf{67.69} & \textbf{51.66} & \textbf{59.60} \\

\bottomrule
\end{tabular}
\end{table}


\subsection{Experimental Setup}

\subsubsection{Evaluation Metrics}
\label{metrics}

We use four core metrics to evaluate detection and classification. Metrics specific to the external benchmarks are introduced together with the corresponding evaluation settings below. For fine-grained hallucination detection, we adopt and refine $\mathbf{F1}_{\mathrm{M}}$ and $\mathbf{F1}_{\mathrm{IoU}}$ from MHALO~\cite{cai2025mhalo} by introducing positional consistency constraints. For hallucination classification, we use $\mathbf{F1}_{\mathrm{Macro}}$ and $\mathbf{F1}_{\mathrm{Micro}}$ scores.

\noindent\textbf{Partial Match Metric $\mathbf{F1}_{\mathrm{M}}$.} To measure span matching quality, we adopt the recall-based partial match score ($\mathrm{PM}_{R}$) proposed by Jafari et al.~\cite{jafari2024target}. However, the original $\mathrm{PM}_{R}$ only considers token overlap, which may produce incorrect matches when identical tokens appear at different positions. We therefore introduce a \emph{positional consistency constraint}: a partial match is valid only when the predicted and ground-truth spans overlap in their positional boundaries. The refined $\mathrm{PM}_{R}$ is defined as:
\begin{equation}
\mathrm{PM}_{R}(g_j)=
\begin{cases}
1, & \exists\, p_i \in P,\ p_i = g_j, \\[4pt]
\dfrac{|T(p_i)\cap T(g_j)|}{|T(g_j)|}, &
\exists\, p_i\in P,\ \max(s_i^p,s_j^g)\!\le\!\min(e_i^p,e_j^g),
\\[4pt]
0, & \text{otherwise}.
\end{cases}
\label{eq:pmr}
\end{equation}
\noindent If multiple predicted spans $p_i$ overlap with $g_j$, we take the one with the highest match score. The precision-based score $\mathrm{PM}_{P}$ is defined analogously. The overall recall and precision are $\mathrm{Rec}_{M} = \frac{1}{m}\sum_{j}\mathrm{PM}_{R}(g_j)$ and $\mathrm{Prec}_{M} = \frac{1}{n}\sum_{i}\mathrm{PM}_{P}(p_i)$, and $\mathbf{F1}_{\mathrm{M}}$ is their harmonic mean.

\noindent\textbf{Token-aware IoU Metric $\mathbf{F1}_{\mathrm{IoU}}$.} While $\mathbf{F1}_{\mathrm{M}}$ captures inclusion relationships between spans, it does not quantify the content-level overlap. We refine the $\mathbf{F1}_{\mathrm{IoU}}$ metric from MHALO by incorporating token-content consistency into the IoU computation. The token-aware IoU is computed only when positional overlap exists:
\begin{equation}
\mathrm{IoU}_{\mathrm{t}}(p_i,g_j)
=
\begin{cases}
\dfrac{|T(p_i)\cap T(g_j)|}{|T(p_i)\cup T(g_j)|},
& \max(s_i^p,s_j^g)\le \min(e_i^p,e_j^g), \\[8pt]
0, & \text{otherwise}.
\end{cases}
\label{eq:iou_t}
\end{equation}
This penalizes spans that share boundaries but have little token overlap. A match is valid only when $\mathrm{IoU}_{\mathrm{t}}(p_i,g_j) \ge \delta$ ($\delta=0.5$), and the optimal one-to-one matching is obtained via the Hungarian algorithm~\cite{kuhn1955hungarian}:
\begin{equation}
\hat{M}
=
\operatorname*{arg\,max}_{M\in\mathcal{M}}
\sum_{(i,j)\in M}
\mathbf{1}\!\left(\mathrm{IoU}_{\mathrm{t}}(p_i,g_j)\ge \delta\right),
\label{eq:iou_matching}
\end{equation}
where $\mathcal{M}$ denotes the set of all one-to-one bipartite matchings between $P$ and $G$:
\begin{equation}
\mathcal{M}
=
\left\{
M\subseteq P\times G
\ \middle|\
\forall (p_i,g_j),(p_{i'},g_{j'})\in M,\ i\neq i',\ j\neq j'
\right\}.
\label{eq:bipartite}
\end{equation}
The final score is $\mathbf{F1}_{\mathrm{IoU}} = 2|\hat{M}|/(|P|+|G|)$, which balances the proportion of correctly matched spans in both predicted and ground-truth sets, providing a stricter evaluation than $\mathbf{F1}_{\mathrm{M}}$ by requiring sufficient token overlap rather than partial inclusion.

\noindent\textbf{Classification Metrics $\mathbf{F1}_{\mathrm{Macro}}$ and $\mathbf{F1}_{\mathrm{Micro}}$.} These metrics evaluate classification across the $C$ hallucination types based on whether each predicted type matches the ground truth. $\mathbf{F1}_{\mathrm{Macro}}$ averages the per-type F1 equally, reflecting balanced performance across rare and common types, while $\mathbf{F1}_{\mathrm{Micro}}$ aggregates TP, FP, and FN globally, reflecting overall classification performance.

\subsubsection{Benchmark} 

Since hallucination diagnosis comprises multiple components, we evaluate detection and classification directly, and assess the practical utility of diagnostic explanations through downstream hallucination mitigation. Across all stages, we deduplicate every evaluation benchmark against our training data at the image, question, and answer levels to ensure no overlap. For the \textbf{detection} stage, we adopt the MHALO benchmark~\cite{cai2025mhalo}, which contains 2,000 samples from four public subsets: RLHF-V~\cite{yu2024rlhf}, M-HalDetect~\cite{gunjal2024mhal}, Geo170K~\cite{gao2023g}, and MathV360K~\cite{shi2024math}.

For the \textbf{classification} stage, we construct a fine-grained benchmark of 733 samples spanning our full taxonomy. It is built from MMBench~\cite{liu2024mmbench}, MMStar~\cite{chen2024mmstar}, OCRBench~\cite{liu2024ocrbench}, RealWorldQA~\cite{xai2024grok}, Geo170K~\cite{gao2023g}, MathV360K~\cite{shi2024math}, M-HalDetect~\cite{gunjal2024mhal}, and RLHF-V~\cite{yu2024rlhf}, covering five task categories: general perception, mathematical reasoning, OCR, spatial reasoning, and hallucination detection. We additionally assess out-of-distribution generalization on HalLoc~\cite{park2025halloc}, an independent classification benchmark. On its 3,079-sample Caption subset, we map our 12 types to HalLoc's object, attribute, and relationship categories and follow its protocol to report Detection Coverage (Det) and end-to-end accuracy (E2E), which requires correct span detection and classification.

For the \textbf{explanation} stage, we assess the practical utility of diagnostic feedback through downstream hallucination mitigation on our classification benchmark and AMBER~\cite{wang2023amber}, an independent LLM-free benchmark of 1,004 images. Following the AMBER protocol, we report CHAIR, Cover, Hal, and Cog, which measure object hallucination, object coverage, response-level hallucination, and cognitive hallucination, respectively.

\subsubsection{Baselines}

We compare against three groups of baselines. First, HalluScope-4B/8B are post-trained from Qwen3-VL-4B/8B-Instruct~\cite{bai2025qwen3}, which we include as base models. Second, we compare against representative MLLMs, including Gemini-3-Flash, Gemini-3-Pro~\cite{team2023gemini}, GPT-4o~\cite{hurst2024gpt}, and GPT-5~\cite{singh2025openai} (GPT-5.5 on HalLoc). Third, we include HALODET-4B and HALODET-8B~\cite{cai2025mhalo}, specialized baselines for fine-grained hallucination detection; as their model weights are not publicly available, we retrain both from the same base models following the original method and data. HALODET targets detection only, so it is excluded from the classification benchmark.

\subsubsection{Implementation Details}

For SFT, we apply LoRA ($r{=}8$) for 1 epoch ($\sim$2 hours). For RL, we run GRPO for 250 steps ($K{=}5$, $\lambda{=}0.3$, $\sim$12 hours). Training is conducted on two H20 144GB GPUs and two B20 180GB GPUs.


\begin{table}[t]
\centering
\small
\caption{Out-of-distribution classification results on the HalLoc Caption benchmark (\%). Det denotes Detection Coverage and E2E denotes end-to-end accuracy (jointly detected and correctly classified). Best results are in bold.}
\label{tab:halloc_ood}
\setlength{\tabcolsep}{4pt}
\renewcommand{\arraystretch}{1.12}
\resizebox{\columnwidth}{!}{%
\begin{tabular}{lcccccccc}
\toprule
& \multicolumn{2}{c}{\textbf{Object}} & \multicolumn{2}{c}{\textbf{Attribute}} & \multicolumn{2}{c}{\textbf{Relationship}} & \multicolumn{2}{c}{\textbf{Overall}} \\
\cmidrule(lr){2-3} \cmidrule(lr){4-5} \cmidrule(lr){6-7} \cmidrule(lr){8-9}
\textbf{Model} & Det & E2E & Det & E2E & Det & E2E & Det & E2E \\
\midrule
Qwen3-VL-8B-Instruct & 65.1 & 64.6 & 42.1 & 24.3 & 14.6 & 5.4 & 44.2 & 35.2 \\
GPT-5.5 & 85.4 & 84.4 & 73.2 & 56.0 & 64.2 & 31.5 & 75.7 & 60.9 \\
Gemini-3-Pro & 81.5 & 81.1 & 68.4 & 50.3 & 51.9 & 46.5 & 69.4 & 61.3 \\
\textbf{HalluScope-8B} & \textbf{92.1} & \textbf{90.7} & \textbf{86.8} & \textbf{57.8} & \textbf{87.7} & \textbf{59.8} & \textbf{89.1} & \textbf{71.0} \\
\bottomrule
\end{tabular}%
}
\end{table}


\begin{figure*}[t]
  \centering
  \includegraphics[width=\textwidth]{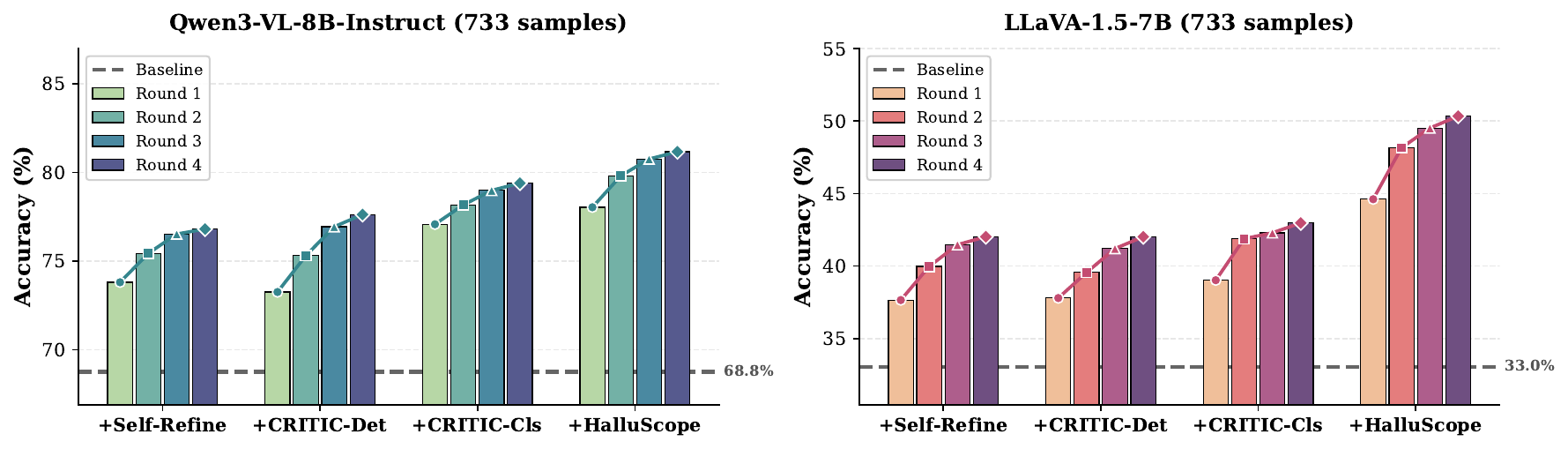}
  \caption{Accuracy (\%) over four feedback rounds with different diagnostic granularities on two target models.}
  \label{fig:mitigation}
  \Description{Dual bar charts showing accuracy improvements over four feedback rounds for Self-Refine, CRITIC-Det, CRITIC-Cls, and HalluScope methods on Qwen3-VL-8B-Instruct (left) and LLaVA-1.5-7B (right).}
\end{figure*}


\subsection{Main Results}

\noindent\textbf{HalluScope achieves state-of-the-art detection on MHALO.} As shown in Table~\ref{tab:results_mhalo}, HalluScope-8B obtains the highest average $\mathbf{F1}_{\mathrm{M}}$ of 64.03 and $\mathbf{F1}_{\mathrm{IoU}}$ of 57.57, surpassing the strongest baseline Gemini-3-Flash by 4.71 and 8.99 points, while HalluScope-4B also beats all baselines. The improvements are most pronounced on the reasoning-oriented Geo170K and MathV360K subsets, where HalluScope-8B reaches $\mathbf{F1}_{\mathrm{M}}$ scores of 74.65 and 79.06, indicating that our joint reward is particularly effective at localizing hallucinations that require mathematical and geometric reasoning. Both models also maintain a near-perfect format validity rate (IF) of at least 99\%. Overall, both models outperform their respective base models and the specialized HALODET baselines, with HalluScope-8B lifting average $\mathbf{F1}_{\mathrm{M}}$ from 36.86 to 64.03 over its base model, confirming the effectiveness of our post-training approach.

\noindent\textbf{Our models achieve the best performance on the hallucination classification benchmark.} As shown in Table~\ref{tab:results_ours}, HalluScope-8B outperforms all baselines across all four metrics, achieving an $\mathbf{F1}_{\mathrm{Macro}}$ of 51.66 and an $\mathbf{F1}_{\mathrm{Micro}}$ of 59.60. Notably, it raises $\mathbf{F1}_{\mathrm{Macro}}$ from 33.73 to 51.66 over its base model, showing that our post-training substantially strengthens the model's classification ability. HalluScope-4B also surpasses Gemini-3-Pro on $\mathbf{F1}_{\mathrm{M}}$, $\mathbf{F1}_{\mathrm{IoU}}$, and $\mathbf{F1}_{\mathrm{Micro}}$, while remaining competitive on $\mathbf{F1}_{\mathrm{Macro}}$. To verify that these gains generalize beyond our data distribution, we further evaluate on the HalLoc benchmark~\cite{park2025halloc}. As shown in Table~\ref{tab:halloc_ood}, HalluScope-8B attains the best E2E in all three categories and an overall E2E of \textbf{71.0}, outperforming GPT-5.5 (60.9) and Gemini-3-Pro (61.3), with an especially large advantage on the harder attribute and relationship categories, confirming that its classification ability generalizes to out-of-distribution data.


\begin{table}[t]
\centering
\footnotesize
\caption{Ablation of the joint reward (HalluScope-8B). \textbf{w/o} drops the classification reward on MHALO and the detection reward on our benchmark. $M_1$/$M_2$ are $\mathbf{F1}_{\mathrm{M}}$/$\mathbf{F1}_{\mathrm{IoU}}$ for MHALO and $\mathbf{F1}_{\mathrm{Macro}}$/$\mathbf{F1}_{\mathrm{Micro}}$ for ours. Best results are in bold.}
\label{tab:ablation}
\setlength{\tabcolsep}{5pt}
\renewcommand{\arraystretch}{1.1}
\resizebox{\columnwidth}{!}{%
\begin{tabular}{llcccc}
\toprule
\multirow{2}{*}{\textbf{Benchmark}} & \multirow{2}{*}{\textbf{Subset}} & \multicolumn{2}{c}{\textbf{Full}} & \multicolumn{2}{c}{\textbf{w/o}} \\
\cmidrule(lr){3-4} \cmidrule(lr){5-6}
 & & $M_1$ & $M_2$ & $M_1$ & $M_2$ \\
\midrule
\multirow{5}{*}{\textbf{MHALO}}
 & RLHF-V      & 55.17 & 43.66 & 53.34 & 44.25 \\
 & M-HalDetect & 47.24 & 45.31 & 48.50 & 45.03 \\
 & Geo170K     & 74.65 & 67.05 & 65.22 & 58.10 \\
 & MathV360K   & 79.06 & 74.26 & 78.07 & 73.82 \\
 & \textit{Average} & \textbf{64.03} & \textbf{57.57} & 61.28 & 55.30 \\
\midrule
\multirow{9}{*}{\textbf{Ours}}
 & MMBench     & 59.16 & 68.97 & 48.14 & 52.22 \\
 & MMStar      & 58.26 & 62.15 & 46.81 & 46.05 \\
 & OCRBench    & 81.60 & 81.02 & 45.68 & 56.18 \\
 & RealWorldQA & 54.82 & 69.21 & 45.31 & 47.23 \\
 & Geo170K     & 28.53 & 41.59 & 17.20 & 23.80 \\
 & MathV360K   & 50.00 & 59.11 & 38.75 & 43.90 \\
 & M-HalDetect & 39.05 & 63.69 & 21.46 & 37.18 \\
 & RLHF-V      & 51.89 & 61.46 & 39.02 & 33.45 \\
 & \textit{Average} & \textbf{51.66} & \textbf{59.60} & 38.17 & 40.54 \\
\bottomrule
\end{tabular}%
}
\end{table}


\subsection{Ablation Study}

To validate our multi-granular joint reward, we train two variants, each using only the detection reward $r_{\mathrm{det}}(y)$ or only the classification reward $r_{\mathrm{cls}}(y)$. As shown in the MHALO block of Table~\ref{tab:ablation}, removing the classification reward lowers average $\mathbf{F1}_{\mathrm{M}}$ from 64.03 to 61.28 and $\mathbf{F1}_{\mathrm{IoU}}$ from 57.57 to 55.30, with the largest drop on the reasoning-oriented Geo170K subset. On our classification benchmark, removing the detection reward causes a larger performance drop, with $\mathbf{F1}_{\mathrm{Macro}}$ falling from 51.66 to 38.17 and $\mathbf{F1}_{\mathrm{Micro}}$ from 59.60 to 40.54. These results show that the two objectives are complementary. The classification reward encourages semantically precise span localization, while the detection reward provides the localized spans required for reliable type prediction. The larger degradation in classification further suggests that accurate detection serves as the foundation for type assignment. Overall, neither reward alone is sufficient, confirming that our multi-granular joint reward is essential for optimizing detection and classification simultaneously.


\begin{figure*}[t]
  \centering
  \includegraphics[width=\textwidth]{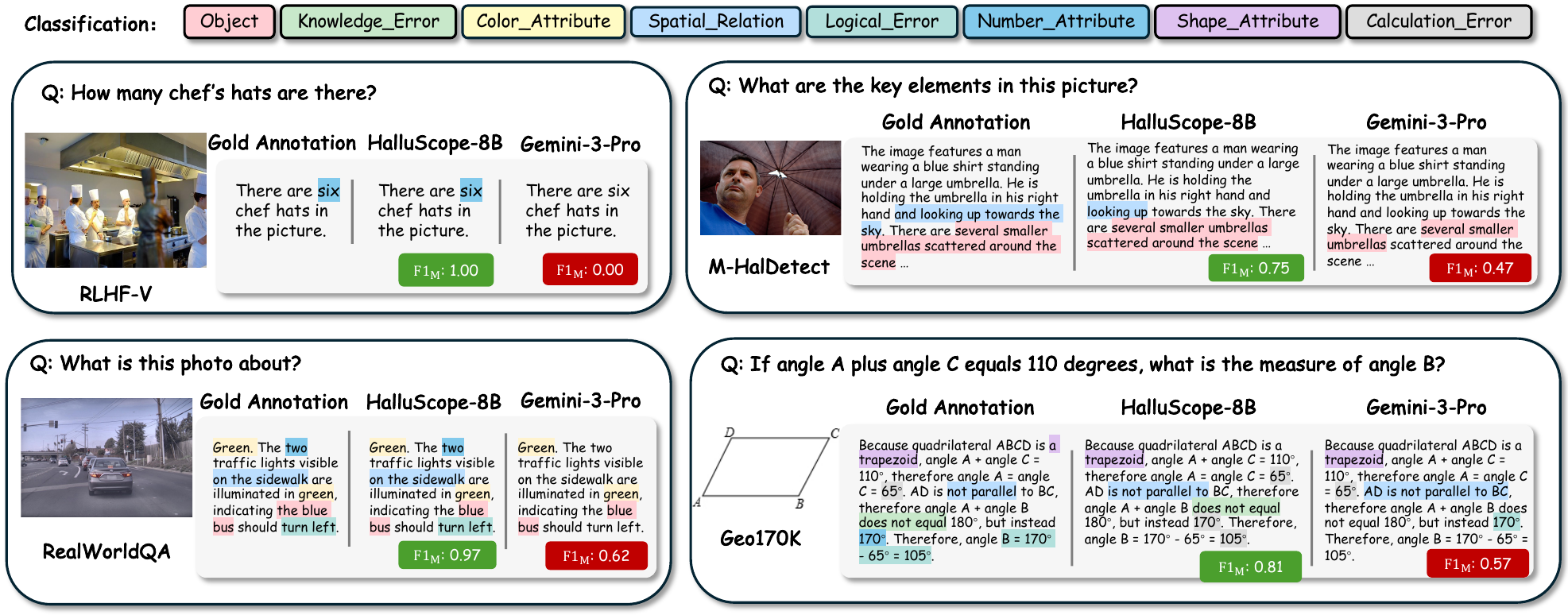}
  \caption{Qualitative comparison between HalluScope-8B and Gemini-3-Pro on four representative samples. Colored spans indicate detected hallucinations with type-specific annotations. $\mathbf{F1}_{\mathrm{M}}$ scores are shown for each prediction.}
  \label{fig:case_study}
  \Description{Four case studies comparing hallucination diagnosis outputs of HalluScope-8B and Gemini-3-Pro across RLHF-V, M-HalDetect, RealWorldQA, and Geo170K datasets.}
\end{figure*}


\begin{table}[t]
\centering
\small
\caption{Diagnosis-driven feedback results on the AMBER benchmark after four rounds. Best results are in bold.}
\label{tab:amber_feedback}
\setlength{\tabcolsep}{2.6pt}
\renewcommand{\arraystretch}{1.12}
\resizebox{\columnwidth}{!}{%
\begin{tabular}{lcccc|cccc}
\toprule
& \multicolumn{4}{c|}{\textbf{Qwen3-VL-8B-Instruct}} & \multicolumn{4}{c}{\textbf{LLaVA-1.5-7B}} \\
\cmidrule(lr){2-5} \cmidrule(lr){6-9}
\textbf{Method} & CHAIR$\downarrow$ & Cover$\uparrow$ & Hal$\downarrow$ & Cog$\downarrow$ & CHAIR$\downarrow$ & Cover$\uparrow$ & Hal$\downarrow$ & Cog$\downarrow$ \\
\midrule
Baseline & 7.5 & 73.7 & 54.2 & 2.7 & 7.5 & 50.7 & 33.3 & 3.9 \\
Self-Refine & 8.8 & \textbf{75.0} & 64.4 & 4.0 & 7.5 & 50.7 & 33.3 & 3.9 \\
CRITIC-Det & 7.6 & 72.8 & 53.1 & 2.9 & 6.6 & 50.2 & 30.4 & 3.3 \\
CRITIC-Cls & 7.6 & 72.9 & 54.2 & 3.0 & 6.2 & 50.0 & 28.3 & 3.1 \\
\textbf{HalluScope} & \textbf{7.4} & 73.4 & \textbf{52.9} & \textbf{2.7} & \textbf{4.2} & \textbf{50.8} & \textbf{20.5} & \textbf{2.0} \\
\bottomrule
\end{tabular}%
}
\end{table}


\subsection{Diagnosis-driven Feedback Experiments}

After evaluating the performance of our models, a natural question arises: what practical value does it provide? The most direct answer is that it can offer diagnostic explanations for incorrect answers, enabling self-improvement of the target model. To this end, we explore a diagnosis-driven multi-round feedback experiment on two models: Qwen3-VL-8B-Instruct and LLaVA-1.5-7B. In each round, diagnostic feedback on the target model's previous response is provided to guide its regeneration.

To examine how diagnostic granularity affects mitigation, we compare four feedback modes. \textbf{Self-Refine}~\cite{madaan2023self} provides only a generic self-correction instruction indicating that the previous response may be incorrect. Following the verify-then-correct paradigm of \textbf{CRITIC}~\cite{gou2023critic}, \textbf{CRITIC-Det} provides detection-level feedback by marking hallucinated spans and generating structured correction instructions, but without type labels, while \textbf{CRITIC-Cls} additionally assigns a hallucination type to each span (e.g., \texttt{Object}, \texttt{Calculation\_Error}) together with type-specific correction instructions. Finally, \textbf{HalluScope} further generates interpretable explanations that specify what is incorrect, why it is incorrect, and how it can be corrected.

As shown in Figure~\ref{fig:mitigation}, finer-grained diagnostic feedback consistently leads to greater accuracy improvements. On Qwen3-VL-8B-Instruct, HalluScope improves accuracy from 68.76\% to 81.17\% after four rounds, outperforming CRITIC-Cls (79.40\%), CRITIC-Det (77.63\%), and Self-Refine (76.81\%). The same ranking is observed on LLaVA-1.5-7B, where HalluScope increases accuracy from 33.02\% to 50.34\%, demonstrating that the benefit transfers across model architectures. As shown in Table~\ref{tab:amber_feedback}, HalluScope consistently reduces hallucination on AMBER while largely preserving object coverage. In particular, on LLaVA-1.5-7B, it lowers CHAIR from 7.5 to 4.2 and Hal from 33.3 to 20.5. By contrast, generic self-refinement can worsen hallucination, as reflected by the increase in Hal from 54.2 to 64.4 on Qwen3-VL-8B-Instruct.


\subsection{Case Study}

Figure~\ref{fig:case_study} presents four examples comparing HalluScope-8B with Gemini-3-Pro across different datasets and hallucination types. In the RLHF-V example, HalluScope-8B precisely identifies the numerical hallucination ``six'' and achieves a perfect $\mathbf{F1}_{\mathrm{M}}$ of 1.0, while Gemini-3-Pro fails to detect any hallucination ($\mathbf{F1}_{\mathrm{M}}$: 0.0). This case highlights that HalluScope-8B is sensitive to fine-grained factual details that a strong general-purpose model may overlook. For the M-HalDetect sample containing multiple hallucination types (Object, Spatial\_Relation), HalluScope-8B locates most hallucinated spans with an $\mathbf{F1}_{\mathrm{M}}$ of 0.75, whereas Gemini-3-Pro only achieves 0.47 with less precise boundaries. The gap here mainly stems from boundary precision: Gemini-3-Pro tends to over-extend spans to neighboring tokens, while HalluScope-8B produces tighter spans that align with the annotated hallucinations, reflecting the effect of our span-level detection reward. The RealWorldQA case demonstrates that HalluScope-8B can accurately detect subtle perceptual errors in real-world scenes ($\mathbf{F1}_{\mathrm{M}}$: 0.97 vs.\ 0.62), where the hallucination concerns easily overlooked visual attributes that require careful grounding in the image. The Geo170K example involves reasoning hallucinations (Logical\_Error, Calculation\_Error, Knowledge\_Error) in a geometry problem, where HalluScope-8B achieves an $\mathbf{F1}_{\mathrm{M}}$ of 0.81 versus 0.57 for Gemini-3-Pro, demonstrating a stronger capability in identifying logical and computational errors that are entangled within a multi-step derivation. Notably, this case shows that HalluScope-8B not only localizes the erroneous reasoning steps but also correctly assigns fine-grained labels, distinguishing logical errors from calculation errors. Across these cases, HalluScope-8B consistently produces more accurate span-level detection and more precise type classification for both perception-level hallucinations, which require faithful grounding in visual evidence, and reasoning-level hallucinations, which require tracing errors through multi-step inference, indicating that the diagnostic ability learned from HalluScope-30K generalizes across heterogeneous tasks and error mechanisms.

\section{CONCLUSION}

In this paper, we introduce fine-grained hallucination diagnosis, a unified task that integrates hallucination detection, classification, and interpretable explanation generation for MLLMs. We develop an automated data generation pipeline to construct HalluScope-30K, comprising nearly 30,000 diagnostically annotated samples from eight datasets and five task categories, and based on this dataset we design a multi-granular joint reward and train HalluScope-4B and HalluScope-8B. Experiments show that our models achieve state-of-the-art performance on the MHALO detection benchmark and on both our and the out-of-distribution HalLoc classification benchmarks. We further demonstrate that detection and classification reinforce each other under joint optimization, while finer-grained diagnostic feedback enables more effective hallucination mitigation, with these benefits also observed on the external AMBER benchmark. We hope our dataset, models, and benchmarks will facilitate future research on fine-grained hallucination diagnosis.

\clearpage
\begin{acks}
This work is supported by the Postdoctoral Fellowship Program of CPSF under Grant Number GZC20251085, the China Postdoctoral Science Foundation under Grant Number 2025M781445 and 2026T190404.
\end{acks}

\bibliographystyle{ACM-Reference-Format}
\bibliography{references}


\begin{thebibliography}{57}


\ifx \showCODEN    \undefined \def \showCODEN     #1{\unskip}     \fi
\ifx \showISBNx    \undefined \def \showISBNx     #1{\unskip}     \fi
\ifx \showISBNxiii \undefined \def \showISBNxiii  #1{\unskip}     \fi
\ifx \showISSN     \undefined \def \showISSN      #1{\unskip}     \fi
\ifx \showLCCN     \undefined \def \showLCCN      #1{\unskip}     \fi
\ifx \shownote     \undefined \def \shownote      #1{#1}          \fi
\ifx \showarticletitle \undefined \def \showarticletitle #1{#1}   \fi
\ifx \showURL      \undefined \def \showURL       {\relax}        \fi
\providecommand\bibfield[2]{#2}
\providecommand\bibinfo[2]{#2}
\providecommand\natexlab[1]{#1}
\providecommand\showeprint[2][]{arXiv:#2}

\bibitem[Antol et~al\mbox{.}(2015)]%
        {antol2015vqa}
\bibfield{author}{\bibinfo{person}{Stanislaw Antol}, \bibinfo{person}{Aishwarya
  Agrawal}, \bibinfo{person}{Jiasen Lu}, \bibinfo{person}{Margaret Mitchell},
  \bibinfo{person}{Dhruv Batra}, \bibinfo{person}{C~Lawrence Zitnick}, {and}
  \bibinfo{person}{Devi Parikh}.} \bibinfo{year}{2015}\natexlab{}.
\newblock \showarticletitle{Vqa: Visual question answering}. In
  \bibinfo{booktitle}{\emph{Proceedings of the IEEE international conference on
  computer vision}}. \bibinfo{pages}{2425--2433}.
\newblock


\bibitem[Bai et~al\mbox{.}(2025)]%
        {bai2025qwen3}
\bibfield{author}{\bibinfo{person}{Shuai Bai}, \bibinfo{person}{Yuxuan Cai},
  \bibinfo{person}{Ruizhe Chen}, \bibinfo{person}{Keqin Chen},
  \bibinfo{person}{Xionghui Chen}, \bibinfo{person}{Zesen Cheng},
  \bibinfo{person}{Lianghao Deng}, \bibinfo{person}{Wei Ding},
  \bibinfo{person}{Chang Gao}, \bibinfo{person}{Chunjiang Ge}, {et~al\mbox{.}}}
  \bibinfo{year}{2025}\natexlab{}.
\newblock \showarticletitle{Qwen3-vl technical report}.
\newblock \bibinfo{journal}{\emph{arXiv preprint arXiv:2511.21631}}
  (\bibinfo{year}{2025}).
\newblock


\bibitem[Bai et~al\mbox{.}(2024)]%
        {bai2024hallucination}
\bibfield{author}{\bibinfo{person}{Zechen Bai}, \bibinfo{person}{Pichao Wang},
  \bibinfo{person}{Tianjun Xiao}, \bibinfo{person}{Tong He},
  \bibinfo{person}{Zongbo Han}, \bibinfo{person}{Zheng Zhang}, {and}
  \bibinfo{person}{Mike~Zheng Shou}.} \bibinfo{year}{2024}\natexlab{}.
\newblock \showarticletitle{Hallucination of multimodal large language models:
  A survey}.
\newblock \bibinfo{journal}{\emph{arXiv preprint arXiv:2404.18930}}
  (\bibinfo{year}{2024}).
\newblock


\bibitem[Cai et~al\mbox{.}(2025)]%
        {cai2025mhalo}
\bibfield{author}{\bibinfo{person}{Yishuo Cai}, \bibinfo{person}{Renjie Gu},
  \bibinfo{person}{Jiaxu Li}, \bibinfo{person}{Xuancheng Huang},
  \bibinfo{person}{Junzhe Chen}, \bibinfo{person}{Xiaotao Gu}, {and}
  \bibinfo{person}{Minlie Huang}.} \bibinfo{year}{2025}\natexlab{}.
\newblock \showarticletitle{MHALO: Evaluating MLLMs as Fine-grained
  Hallucination Detectors}. In \bibinfo{booktitle}{\emph{Findings of the
  Association for Computational Linguistics: ACL 2025}}.
  \bibinfo{pages}{9197--9222}.
\newblock


\bibitem[Chen et~al\mbox{.}(2024a)]%
        {chen2024mmstar}
\bibfield{author}{\bibinfo{person}{Lin Chen}, \bibinfo{person}{Jinsong Li},
  \bibinfo{person}{Xiaoyi Dong}, \bibinfo{person}{Pan Zhang},
  \bibinfo{person}{Yuhang Zang}, \bibinfo{person}{Zehui Chen},
  \bibinfo{person}{Haodong Duan}, \bibinfo{person}{Jiaqi Wang},
  \bibinfo{person}{Yu Qiao}, \bibinfo{person}{Dahua Lin}, {et~al\mbox{.}}}
  \bibinfo{year}{2024}\natexlab{a}.
\newblock \showarticletitle{Are we on the right way for evaluating large
  vision-language models?}
\newblock \bibinfo{journal}{\emph{Advances in Neural Information Processing
  Systems}}  \bibinfo{volume}{37} (\bibinfo{year}{2024}),
  \bibinfo{pages}{27056--27087}.
\newblock


\bibitem[Chen et~al\mbox{.}(2015)]%
        {chen2015microsoft}
\bibfield{author}{\bibinfo{person}{Xinlei Chen}, \bibinfo{person}{Hao Fang},
  \bibinfo{person}{Tsung-Yi Lin}, \bibinfo{person}{Ramakrishna Vedantam},
  \bibinfo{person}{Saurabh Gupta}, \bibinfo{person}{Piotr Doll{\'a}r}, {and}
  \bibinfo{person}{C~Lawrence Zitnick}.} \bibinfo{year}{2015}\natexlab{}.
\newblock \showarticletitle{Microsoft coco captions: Data collection and
  evaluation server}.
\newblock \bibinfo{journal}{\emph{arXiv preprint arXiv:1504.00325}}
  (\bibinfo{year}{2015}).
\newblock


\bibitem[Chen et~al\mbox{.}(2024b)]%
        {chen2024unified}
\bibfield{author}{\bibinfo{person}{Xiang Chen}, \bibinfo{person}{Chenxi Wang},
  \bibinfo{person}{Yida Xue}, \bibinfo{person}{Ningyu Zhang},
  \bibinfo{person}{Xiaoyan Yang}, \bibinfo{person}{Qiang Li},
  \bibinfo{person}{Yue Shen}, \bibinfo{person}{Lei Liang},
  \bibinfo{person}{Jinjie Gu}, {and} \bibinfo{person}{Huajun Chen}.}
  \bibinfo{year}{2024}\natexlab{b}.
\newblock \showarticletitle{Unified hallucination detection for multimodal
  large language models}. In \bibinfo{booktitle}{\emph{Proceedings of the 62nd
  Annual Meeting of the Association for Computational Linguistics (Volume 1:
  Long Papers)}}. \bibinfo{pages}{3235--3252}.
\newblock


\bibitem[Chen et~al\mbox{.}(2026)]%
        {chen2026survey}
\bibfield{author}{\bibinfo{person}{Zhiyuan Chen}, \bibinfo{person}{Yuecong
  Min}, \bibinfo{person}{Jie Zhang}, \bibinfo{person}{Bei Yan},
  \bibinfo{person}{Jiahao Wang}, \bibinfo{person}{Xiaozhen Wang}, {and}
  \bibinfo{person}{Shiguang Shan}.} \bibinfo{year}{2026}\natexlab{}.
\newblock \showarticletitle{A survey of multimodal hallucination evaluation and
  detection}.
\newblock \bibinfo{journal}{\emph{International Journal of Computer Vision}}
  \bibinfo{volume}{134}, \bibinfo{number}{3} (\bibinfo{year}{2026}),
  \bibinfo{pages}{131}.
\newblock


\bibitem[Chu et~al\mbox{.}(2025)]%
        {chu2025reducing}
\bibfield{author}{\bibinfo{person}{Yun-Wei Chu}, \bibinfo{person}{Kai Zhang},
  \bibinfo{person}{Christopher Malon}, {and} \bibinfo{person}{Martin~Renqiang
  Min}.} \bibinfo{year}{2025}\natexlab{}.
\newblock \showarticletitle{Reducing hallucinations of medical multimodal large
  language models with visual retrieval-augmented generation}.
\newblock \bibinfo{journal}{\emph{arXiv preprint arXiv:2502.15040}}
  (\bibinfo{year}{2025}).
\newblock


\bibitem[Fieback et~al\mbox{.}(2025)]%
        {fieback2025ecd}
\bibfield{author}{\bibinfo{person}{Laura Fieback}, \bibinfo{person}{Nishilkumar
  Balar}, \bibinfo{person}{Jakob Spiegelberg}, {and} \bibinfo{person}{Hanno
  Gottschalk}.} \bibinfo{year}{2025}\natexlab{}.
\newblock \showarticletitle{Efficient Contrastive Decoding with Probabilistic
  Hallucination Detection-Mitigating Hallucinations in Large Vision Language
  Models}.
\newblock \bibinfo{journal}{\emph{arXiv preprint arXiv:2504.12137}}
  (\bibinfo{year}{2025}).
\newblock


\bibitem[Fieback et~al\mbox{.}(2024)]%
        {fieback2024metatoken}
\bibfield{author}{\bibinfo{person}{Laura Fieback}, \bibinfo{person}{Jakob
  Spiegelberg}, {and} \bibinfo{person}{Hanno Gottschalk}.}
  \bibinfo{year}{2024}\natexlab{}.
\newblock \showarticletitle{Metatoken: Detecting hallucination in image
  descriptions by meta classification}.
\newblock \bibinfo{journal}{\emph{arXiv preprint arXiv:2405.19186}}
  (\bibinfo{year}{2024}).
\newblock


\bibitem[Fu et~al\mbox{.}(2024)]%
        {fu2024tldr}
\bibfield{author}{\bibinfo{person}{Deqing Fu}, \bibinfo{person}{Tong Xiao},
  \bibinfo{person}{Rui Wang}, \bibinfo{person}{Wang Zhu},
  \bibinfo{person}{Pengchuan Zhang}, \bibinfo{person}{Guan Pang},
  \bibinfo{person}{Robin Jia}, {and} \bibinfo{person}{Lawrence Chen}.}
  \bibinfo{year}{2024}\natexlab{}.
\newblock \showarticletitle{Tldr: Token-level detective reward model for large
  vision language models}.
\newblock \bibinfo{journal}{\emph{arXiv preprint arXiv:2410.04734}}
  (\bibinfo{year}{2024}).
\newblock


\bibitem[Gao et~al\mbox{.}(2023)]%
        {gao2023g}
\bibfield{author}{\bibinfo{person}{Jiahui Gao}, \bibinfo{person}{Renjie Pi},
  \bibinfo{person}{Jipeng Zhang}, \bibinfo{person}{Jiacheng Ye},
  \bibinfo{person}{Wanjun Zhong}, \bibinfo{person}{Yufei Wang},
  \bibinfo{person}{Lanqing Hong}, \bibinfo{person}{Jianhua Han},
  \bibinfo{person}{Hang Xu}, \bibinfo{person}{Zhenguo Li}, {et~al\mbox{.}}}
  \bibinfo{year}{2023}\natexlab{}.
\newblock \showarticletitle{G-llava: Solving geometric problem with multi-modal
  large language model}.
\newblock \bibinfo{journal}{\emph{arXiv preprint arXiv:2312.11370}}
  (\bibinfo{year}{2023}).
\newblock


\bibitem[Glm et~al\mbox{.}(2024)]%
        {glm2024chatglm}
\bibfield{author}{\bibinfo{person}{Team Glm}, \bibinfo{person}{Aohan Zeng},
  \bibinfo{person}{Bin Xu}, \bibinfo{person}{Bowen Wang},
  \bibinfo{person}{Chenhui Zhang}, \bibinfo{person}{Da Yin},
  \bibinfo{person}{Dan Zhang}, \bibinfo{person}{Diego Rojas},
  \bibinfo{person}{Guanyu Feng}, \bibinfo{person}{Hanlin Zhao},
  {et~al\mbox{.}}} \bibinfo{year}{2024}\natexlab{}.
\newblock \showarticletitle{Chatglm: A family of large language models from
  glm-130b to glm-4 all tools}.
\newblock \bibinfo{journal}{\emph{arXiv preprint arXiv:2406.12793}}
  (\bibinfo{year}{2024}).
\newblock


\bibitem[Gou et~al\mbox{.}(2023)]%
        {gou2023critic}
\bibfield{author}{\bibinfo{person}{Zhibin Gou}, \bibinfo{person}{Zhihong Shao},
  \bibinfo{person}{Yeyun Gong}, \bibinfo{person}{Yelong Shen},
  \bibinfo{person}{Yujiu Yang}, \bibinfo{person}{Nan Duan}, {and}
  \bibinfo{person}{Weizhu Chen}.} \bibinfo{year}{2023}\natexlab{}.
\newblock \showarticletitle{Critic: Large language models can self-correct with
  tool-interactive critiquing}.
\newblock \bibinfo{journal}{\emph{arXiv preprint arXiv:2305.11738}}
  (\bibinfo{year}{2023}).
\newblock


\bibitem[Gunjal et~al\mbox{.}(2024)]%
        {gunjal2024mhal}
\bibfield{author}{\bibinfo{person}{Anisha Gunjal}, \bibinfo{person}{Jihan Yin},
  {and} \bibinfo{person}{Erhan Bas}.} \bibinfo{year}{2024}\natexlab{}.
\newblock \showarticletitle{Detecting and preventing hallucinations in large
  vision language models}. In \bibinfo{booktitle}{\emph{Proceedings of the AAAI
  Conference on Artificial Intelligence}}, Vol.~\bibinfo{volume}{38}.
  \bibinfo{pages}{18135--18143}.
\newblock


\bibitem[Guo et~al\mbox{.}(2025)]%
        {guo2025deepseek}
\bibfield{author}{\bibinfo{person}{Daya Guo}, \bibinfo{person}{Dejian Yang},
  \bibinfo{person}{Haowei Zhang}, \bibinfo{person}{Junxiao Song},
  \bibinfo{person}{Peiyi Wang}, \bibinfo{person}{Qihao Zhu},
  \bibinfo{person}{Runxin Xu}, \bibinfo{person}{Ruoyu Zhang},
  \bibinfo{person}{Shirong Ma}, \bibinfo{person}{Xiao Bi}, {et~al\mbox{.}}}
  \bibinfo{year}{2025}\natexlab{}.
\newblock \showarticletitle{DeepSeek-R1 incentivizes reasoning in LLMs through
  reinforcement learning}.
\newblock \bibinfo{journal}{\emph{Nature}} \bibinfo{volume}{645},
  \bibinfo{number}{8081} (\bibinfo{year}{2025}), \bibinfo{pages}{633--638}.
\newblock


\bibitem[Hu et~al\mbox{.}(2022)]%
        {hu2022lora}
\bibfield{author}{\bibinfo{person}{Edward~J Hu}, \bibinfo{person}{Yelong Shen},
  \bibinfo{person}{Phillip Wallis}, \bibinfo{person}{Zeyuan Allen-Zhu},
  \bibinfo{person}{Yuanzhi Li}, \bibinfo{person}{Shean Wang},
  \bibinfo{person}{Liang Wang}, \bibinfo{person}{Weizhu Chen}, {et~al\mbox{.}}}
  \bibinfo{year}{2022}\natexlab{}.
\newblock \showarticletitle{Lora: Low-rank adaptation of large language
  models.}
\newblock \bibinfo{journal}{\emph{Iclr}} \bibinfo{volume}{1},
  \bibinfo{number}{2} (\bibinfo{year}{2022}), \bibinfo{pages}{3}.
\newblock


\bibitem[Hurst et~al\mbox{.}(2024)]%
        {hurst2024gpt}
\bibfield{author}{\bibinfo{person}{Aaron Hurst}, \bibinfo{person}{Adam Lerer},
  \bibinfo{person}{Adam~P Goucher}, \bibinfo{person}{Adam Perelman},
  \bibinfo{person}{Aditya Ramesh}, \bibinfo{person}{Aidan Clark},
  \bibinfo{person}{AJ Ostrow}, \bibinfo{person}{Akila Welihinda},
  \bibinfo{person}{Alan Hayes}, \bibinfo{person}{Alec Radford},
  {et~al\mbox{.}}} \bibinfo{year}{2024}\natexlab{}.
\newblock \showarticletitle{Gpt-4o system card}.
\newblock \bibinfo{journal}{\emph{arXiv preprint arXiv:2410.21276}}
  (\bibinfo{year}{2024}).
\newblock


\bibitem[Jafari et~al\mbox{.}(2024)]%
        {jafari2024target}
\bibfield{author}{\bibinfo{person}{Nazanin Jafari}, \bibinfo{person}{James
  Allan}, {and} \bibinfo{person}{Sheikh~Muhammad Sarwar}.}
  \bibinfo{year}{2024}\natexlab{}.
\newblock \showarticletitle{Target span detection for implicit harmful
  content}. In \bibinfo{booktitle}{\emph{Proceedings of the 2024 ACM SIGIR
  International Conference on Theory of Information Retrieval}}.
  \bibinfo{pages}{117--122}.
\newblock


\bibitem[Kuang et~al\mbox{.}(2025)]%
        {kuang2025natural}
\bibfield{author}{\bibinfo{person}{Jiayi Kuang}, \bibinfo{person}{Ying Shen},
  \bibinfo{person}{Jingyou Xie}, \bibinfo{person}{Haohao Luo},
  \bibinfo{person}{Zhe Xu}, \bibinfo{person}{Ronghao Li},
  \bibinfo{person}{Yinghui Li}, \bibinfo{person}{Xianfeng Cheng},
  \bibinfo{person}{Xika Lin}, {and} \bibinfo{person}{Yu Han}.}
  \bibinfo{year}{2025}\natexlab{}.
\newblock \showarticletitle{Natural language understanding and inference with
  mllm in visual question answering: A survey}.
\newblock \bibinfo{journal}{\emph{Comput. Surveys}} \bibinfo{volume}{57},
  \bibinfo{number}{8} (\bibinfo{year}{2025}), \bibinfo{pages}{1--36}.
\newblock


\bibitem[Kuhn(1955)]%
        {kuhn1955hungarian}
\bibfield{author}{\bibinfo{person}{Harold~W Kuhn}.}
  \bibinfo{year}{1955}\natexlab{}.
\newblock \showarticletitle{The Hungarian method for the assignment problem}.
\newblock \bibinfo{journal}{\emph{Naval research logistics quarterly}}
  \bibinfo{volume}{2}, \bibinfo{number}{1-2} (\bibinfo{year}{1955}),
  \bibinfo{pages}{83--97}.
\newblock


\bibitem[Leng et~al\mbox{.}(2024)]%
        {leng2024mitigating}
\bibfield{author}{\bibinfo{person}{Sicong Leng}, \bibinfo{person}{Hang Zhang},
  \bibinfo{person}{Guanzheng Chen}, \bibinfo{person}{Xin Li},
  \bibinfo{person}{Shijian Lu}, \bibinfo{person}{Chunyan Miao}, {and}
  \bibinfo{person}{Lidong Bing}.} \bibinfo{year}{2024}\natexlab{}.
\newblock \showarticletitle{Mitigating object hallucinations in large
  vision-language models through visual contrastive decoding}. In
  \bibinfo{booktitle}{\emph{Proceedings of the IEEE/CVF Conference on Computer
  Vision and Pattern Recognition}}. \bibinfo{pages}{13872--13882}.
\newblock


\bibitem[Li et~al\mbox{.}(2023b)]%
        {li2023llavamed}
\bibfield{author}{\bibinfo{person}{Chunyuan Li}, \bibinfo{person}{Cliff Wong},
  \bibinfo{person}{Sheng Zhang}, \bibinfo{person}{Naoto Usuyama},
  \bibinfo{person}{Haotian Liu}, \bibinfo{person}{Jianwei Yang},
  \bibinfo{person}{Tristan Naumann}, \bibinfo{person}{Hoifung Poon}, {and}
  \bibinfo{person}{Jianfeng Gao}.} \bibinfo{year}{2023}\natexlab{b}.
\newblock \showarticletitle{Llava-med: Training a large language-and-vision
  assistant for biomedicine in one day}.
\newblock \bibinfo{journal}{\emph{Advances in Neural Information Processing
  Systems}}  \bibinfo{volume}{36} (\bibinfo{year}{2023}),
  \bibinfo{pages}{28541--28564}.
\newblock


\bibitem[Li et~al\mbox{.}(2023a)]%
        {li2023blip}
\bibfield{author}{\bibinfo{person}{Junnan Li}, \bibinfo{person}{Dongxu Li},
  \bibinfo{person}{Silvio Savarese}, {and} \bibinfo{person}{Steven Hoi}.}
  \bibinfo{year}{2023}\natexlab{a}.
\newblock \showarticletitle{Blip-2: Bootstrapping language-image pre-training
  with frozen image encoders and large language models}. In
  \bibinfo{booktitle}{\emph{International conference on machine learning}}.
  PMLR, \bibinfo{pages}{19730--19742}.
\newblock


\bibitem[Liu et~al\mbox{.}(2024c)]%
        {liu2024survey}
\bibfield{author}{\bibinfo{person}{Hanchao Liu}, \bibinfo{person}{Wenyuan Xue},
  \bibinfo{person}{Yifei Chen}, \bibinfo{person}{Dapeng Chen},
  \bibinfo{person}{Xiutian Zhao}, \bibinfo{person}{Ke Wang},
  \bibinfo{person}{Liping Hou}, \bibinfo{person}{Rongjun Li}, {and}
  \bibinfo{person}{Wei Peng}.} \bibinfo{year}{2024}\natexlab{c}.
\newblock \showarticletitle{A survey on hallucination in large vision-language
  models}.
\newblock \bibinfo{journal}{\emph{arXiv preprint arXiv:2402.00253}}
  (\bibinfo{year}{2024}).
\newblock


\bibitem[Liu et~al\mbox{.}(2025)]%
        {liu2025reducing}
\bibfield{author}{\bibinfo{person}{Sheng Liu}, \bibinfo{person}{Haotian Ye},
  {and} \bibinfo{person}{James Zou}.} \bibinfo{year}{2025}\natexlab{}.
\newblock \showarticletitle{Reducing hallucinations in large vision-language
  models via latent space steering}. In \bibinfo{booktitle}{\emph{The
  Thirteenth International Conference on Learning Representations}}.
\newblock


\bibitem[Liu et~al\mbox{.}(2024a)]%
        {liu2024mmbench}
\bibfield{author}{\bibinfo{person}{Yuan Liu}, \bibinfo{person}{Haodong Duan},
  \bibinfo{person}{Yuanhan Zhang}, \bibinfo{person}{Bo Li},
  \bibinfo{person}{Songyang Zhang}, \bibinfo{person}{Wangbo Zhao},
  \bibinfo{person}{Yike Yuan}, \bibinfo{person}{Jiaqi Wang},
  \bibinfo{person}{Conghui He}, \bibinfo{person}{Ziwei Liu}, {et~al\mbox{.}}}
  \bibinfo{year}{2024}\natexlab{a}.
\newblock \showarticletitle{Mmbench: Is your multi-modal model an all-around
  player?}. In \bibinfo{booktitle}{\emph{European conference on computer
  vision}}. Springer, \bibinfo{pages}{216--233}.
\newblock


\bibitem[Liu et~al\mbox{.}(2024b)]%
        {liu2024ocrbench}
\bibfield{author}{\bibinfo{person}{Yuliang Liu}, \bibinfo{person}{Zhang Li},
  \bibinfo{person}{Mingxin Huang}, \bibinfo{person}{Biao Yang},
  \bibinfo{person}{Wenwen Yu}, \bibinfo{person}{Chunyuan Li},
  \bibinfo{person}{Xu-Cheng Yin}, \bibinfo{person}{Cheng-Lin Liu},
  \bibinfo{person}{Lianwen Jin}, {and} \bibinfo{person}{Xiang Bai}.}
  \bibinfo{year}{2024}\natexlab{b}.
\newblock \showarticletitle{Ocrbench: on the hidden mystery of ocr in large
  multimodal models}.
\newblock \bibinfo{journal}{\emph{Science China Information Sciences}}
  \bibinfo{volume}{67}, \bibinfo{number}{12} (\bibinfo{year}{2024}),
  \bibinfo{pages}{220102}.
\newblock


\bibitem[Madaan et~al\mbox{.}(2023)]%
        {madaan2023self}
\bibfield{author}{\bibinfo{person}{Aman Madaan}, \bibinfo{person}{Niket
  Tandon}, \bibinfo{person}{Prakhar Gupta}, \bibinfo{person}{Skyler Hallinan},
  \bibinfo{person}{Luyu Gao}, \bibinfo{person}{Sarah Wiegreffe},
  \bibinfo{person}{Uri Alon}, \bibinfo{person}{Nouha Dziri},
  \bibinfo{person}{Shrimai Prabhumoye}, \bibinfo{person}{Yiming Yang},
  {et~al\mbox{.}}} \bibinfo{year}{2023}\natexlab{}.
\newblock \showarticletitle{Self-refine: Iterative refinement with
  self-feedback}.
\newblock \bibinfo{journal}{\emph{Advances in neural information processing
  systems}}  \bibinfo{volume}{36} (\bibinfo{year}{2023}),
  \bibinfo{pages}{46534--46594}.
\newblock


\bibitem[Mishra et~al\mbox{.}(2024)]%
        {mishra2024fine}
\bibfield{author}{\bibinfo{person}{Abhika Mishra}, \bibinfo{person}{Akari
  Asai}, \bibinfo{person}{Vidhisha Balachandran}, \bibinfo{person}{Yizhong
  Wang}, \bibinfo{person}{Graham Neubig}, \bibinfo{person}{Yulia Tsvetkov},
  {and} \bibinfo{person}{Hannaneh Hajishirzi}.}
  \bibinfo{year}{2024}\natexlab{}.
\newblock \showarticletitle{Fine-grained hallucination detection and editing
  for language models}.
\newblock \bibinfo{journal}{\emph{arXiv preprint arXiv:2401.06855}}
  (\bibinfo{year}{2024}).
\newblock


\bibitem[Park et~al\mbox{.}(2025)]%
        {park2025halloc}
\bibfield{author}{\bibinfo{person}{Eunkyu Park}, \bibinfo{person}{Minyeong
  Kim}, {and} \bibinfo{person}{Gunhee Kim}.} \bibinfo{year}{2025}\natexlab{}.
\newblock \showarticletitle{Halloc: Token-level localization of hallucinations
  for vision language models}. In \bibinfo{booktitle}{\emph{Proceedings of the
  IEEE/CVF Conference on Computer Vision and Pattern Recognition}}.
  \bibinfo{pages}{29893--29903}.
\newblock


\bibitem[Peng et~al\mbox{.}(2023)]%
        {peng2023kosmos}
\bibfield{author}{\bibinfo{person}{Zhiliang Peng}, \bibinfo{person}{Wenhui
  Wang}, \bibinfo{person}{Li Dong}, \bibinfo{person}{Yaru Hao},
  \bibinfo{person}{Shaohan Huang}, \bibinfo{person}{Shuming Ma}, {and}
  \bibinfo{person}{Furu Wei}.} \bibinfo{year}{2023}\natexlab{}.
\newblock \showarticletitle{Kosmos-2: Grounding multimodal large language
  models to the world}.
\newblock \bibinfo{journal}{\emph{arXiv preprint arXiv:2306.14824}}
  (\bibinfo{year}{2023}).
\newblock


\bibitem[Sahoo et~al\mbox{.}(2024)]%
        {sahoo2024comprehensive}
\bibfield{author}{\bibinfo{person}{Pranab Sahoo}, \bibinfo{person}{Prabhash
  Meharia}, \bibinfo{person}{Akash Ghosh}, \bibinfo{person}{Sriparna Saha},
  \bibinfo{person}{Vinija Jain}, {and} \bibinfo{person}{Aman Chadha}.}
  \bibinfo{year}{2024}\natexlab{}.
\newblock \showarticletitle{A comprehensive survey of hallucination in large
  language, image, video and audio foundation models}.
\newblock \bibinfo{journal}{\emph{Findings of the Association for Computational
  Linguistics: EMNLP 2024}} (\bibinfo{year}{2024}),
  \bibinfo{pages}{11709--11724}.
\newblock


\bibitem[Sahu et~al\mbox{.}(2024)]%
        {sahu2024pelican}
\bibfield{author}{\bibinfo{person}{Pritish Sahu}, \bibinfo{person}{Karan
  Sikka}, {and} \bibinfo{person}{Ajay Divakaran}.}
  \bibinfo{year}{2024}\natexlab{}.
\newblock \showarticletitle{Pelican: Correcting hallucination in vision-llms
  via claim decomposition and program of thought verification}. In
  \bibinfo{booktitle}{\emph{Proceedings of the 2024 Conference on Empirical
  Methods in Natural Language Processing}}. \bibinfo{pages}{8228--8248}.
\newblock


\bibitem[Shi et~al\mbox{.}(2024)]%
        {shi2024math}
\bibfield{author}{\bibinfo{person}{Wenhao Shi}, \bibinfo{person}{Zhiqiang Hu},
  \bibinfo{person}{Yi Bin}, \bibinfo{person}{Junhua Liu}, \bibinfo{person}{Yang
  Yang}, \bibinfo{person}{See~Kiong Ng}, \bibinfo{person}{Lidong Bing}, {and}
  \bibinfo{person}{Roy Ka-Wei Lee}.} \bibinfo{year}{2024}\natexlab{}.
\newblock \showarticletitle{Math-llava: Bootstrapping mathematical reasoning
  for multimodal large language models}. In \bibinfo{booktitle}{\emph{Findings
  of the Association for Computational Linguistics: EMNLP 2024}}.
  \bibinfo{pages}{4663--4680}.
\newblock


\bibitem[Singh et~al\mbox{.}(2025)]%
        {singh2025openai}
\bibfield{author}{\bibinfo{person}{Aaditya Singh}, \bibinfo{person}{Adam Fry},
  \bibinfo{person}{Adam Perelman}, \bibinfo{person}{Adam Tart},
  \bibinfo{person}{Adi Ganesh}, \bibinfo{person}{Ahmed El-Kishky},
  \bibinfo{person}{Aidan McLaughlin}, \bibinfo{person}{Aiden Low},
  \bibinfo{person}{AJ Ostrow}, \bibinfo{person}{Akhila Ananthram},
  {et~al\mbox{.}}} \bibinfo{year}{2025}\natexlab{}.
\newblock \showarticletitle{Openai gpt-5 system card}.
\newblock \bibinfo{journal}{\emph{arXiv preprint arXiv:2601.03267}}
  (\bibinfo{year}{2025}).
\newblock


\bibitem[Sun et~al\mbox{.}(2024)]%
        {sun2024aligning}
\bibfield{author}{\bibinfo{person}{Zhiqing Sun}, \bibinfo{person}{Sheng Shen},
  \bibinfo{person}{Shengcao Cao}, \bibinfo{person}{Haotian Liu},
  \bibinfo{person}{Chunyuan Li}, \bibinfo{person}{Yikang Shen},
  \bibinfo{person}{Chuang Gan}, \bibinfo{person}{Liangyan Gui},
  \bibinfo{person}{Yu-Xiong Wang}, \bibinfo{person}{Yiming Yang},
  {et~al\mbox{.}}} \bibinfo{year}{2024}\natexlab{}.
\newblock \showarticletitle{Aligning large multimodal models with factually
  augmented rlhf}. In \bibinfo{booktitle}{\emph{Findings of the Association for
  Computational Linguistics: ACL 2024}}. \bibinfo{pages}{13088--13110}.
\newblock


\bibitem[Team et~al\mbox{.}(2023)]%
        {team2023gemini}
\bibfield{author}{\bibinfo{person}{Gemini Team}, \bibinfo{person}{Rohan Anil},
  \bibinfo{person}{Sebastian Borgeaud}, \bibinfo{person}{Jean-Baptiste
  Alayrac}, \bibinfo{person}{Jiahui Yu}, \bibinfo{person}{Radu Soricut},
  \bibinfo{person}{Johan Schalkwyk}, \bibinfo{person}{Andrew~M Dai},
  \bibinfo{person}{Anja Hauth}, \bibinfo{person}{Katie Millican},
  {et~al\mbox{.}}} \bibinfo{year}{2023}\natexlab{}.
\newblock \showarticletitle{Gemini: a family of highly capable multimodal
  models}.
\newblock \bibinfo{journal}{\emph{arXiv preprint arXiv:2312.11805}}
  (\bibinfo{year}{2023}).
\newblock


\bibitem[Wang et~al\mbox{.}(2023a)]%
        {wang2023amber}
\bibfield{author}{\bibinfo{person}{Junyang Wang}, \bibinfo{person}{Yuhang
  Wang}, \bibinfo{person}{Guohai Xu}, \bibinfo{person}{Jing Zhang},
  \bibinfo{person}{Yukai Gu}, \bibinfo{person}{Haitao Jia},
  \bibinfo{person}{Jiaqi Wang}, \bibinfo{person}{Haiyang Xu},
  \bibinfo{person}{Ming Yan}, \bibinfo{person}{Ji Zhang}, {et~al\mbox{.}}}
  \bibinfo{year}{2023}\natexlab{a}.
\newblock \showarticletitle{Amber: An llm-free multi-dimensional benchmark for
  mllms hallucination evaluation}.
\newblock \bibinfo{journal}{\emph{arXiv preprint arXiv:2311.07397}}
  (\bibinfo{year}{2023}).
\newblock


\bibitem[Wang et~al\mbox{.}(2023b)]%
        {wang2023evaluation}
\bibfield{author}{\bibinfo{person}{Junyang Wang}, \bibinfo{person}{Yiyang
  Zhou}, \bibinfo{person}{Guohai Xu}, \bibinfo{person}{Pengcheng Shi},
  \bibinfo{person}{Chenlin Zhao}, \bibinfo{person}{Haiyang Xu},
  \bibinfo{person}{Qinghao Ye}, \bibinfo{person}{Ming Yan}, \bibinfo{person}{Ji
  Zhang}, \bibinfo{person}{Jihua Zhu}, {et~al\mbox{.}}}
  \bibinfo{year}{2023}\natexlab{b}.
\newblock \showarticletitle{Evaluation and analysis of hallucination in large
  vision-language models}.
\newblock \bibinfo{journal}{\emph{arXiv preprint arXiv:2308.15126}}
  (\bibinfo{year}{2023}).
\newblock


\bibitem[Wang et~al\mbox{.}(2024b)]%
        {wang2024mitigating}
\bibfield{author}{\bibinfo{person}{Xintong Wang}, \bibinfo{person}{Jingheng
  Pan}, \bibinfo{person}{Liang Ding}, {and} \bibinfo{person}{Chris Biemann}.}
  \bibinfo{year}{2024}\natexlab{b}.
\newblock \showarticletitle{Mitigating hallucinations in large vision-language
  models with instruction contrastive decoding}. In
  \bibinfo{booktitle}{\emph{Findings of the Association for Computational
  Linguistics: ACL 2024}}. \bibinfo{pages}{15840--15853}.
\newblock


\bibitem[Wang et~al\mbox{.}(2024a)]%
        {wang2024exploring}
\bibfield{author}{\bibinfo{person}{Yiqi Wang}, \bibinfo{person}{Wentao Chen},
  \bibinfo{person}{Xiaotian Han}, \bibinfo{person}{Xudong Lin},
  \bibinfo{person}{Haiteng Zhao}, \bibinfo{person}{Yongfei Liu},
  \bibinfo{person}{Bohan Zhai}, \bibinfo{person}{Jianbo Yuan},
  \bibinfo{person}{Quanzeng You}, {and} \bibinfo{person}{Hongxia Yang}.}
  \bibinfo{year}{2024}\natexlab{a}.
\newblock \showarticletitle{Exploring the reasoning abilities of multimodal
  large language models (mllms): A comprehensive survey on emerging trends in
  multimodal reasoning}.
\newblock \bibinfo{journal}{\emph{arXiv preprint arXiv:2401.06805}}
  (\bibinfo{year}{2024}).
\newblock


\bibitem[Whitehead et~al\mbox{.}(2024)]%
        {whitehead2024pre}
\bibfield{author}{\bibinfo{person}{Spencer Whitehead}, \bibinfo{person}{Jacob
  Phillips}, {and} \bibinfo{person}{Sean Hendryx}.}
  \bibinfo{year}{2024}\natexlab{}.
\newblock \showarticletitle{Pre-Training Multimodal Hallucination Detectors
  with Corrupted Grounding Data}.
\newblock \bibinfo{journal}{\emph{arXiv preprint arXiv:2409.00238}}
  (\bibinfo{year}{2024}).
\newblock


\bibitem[{xAI}(2024)]%
        {xai2024grok}
\bibfield{author}{\bibinfo{person}{{xAI}}.} \bibinfo{year}{2024}\natexlab{}.
\newblock \bibinfo{title}{Grok-1.5V: Multimodal Understanding with Grok}.
\newblock
\urldef\tempurl%
\url{https://x.ai/blog/grok-1.5v}
\showURL{%
\tempurl}


\bibitem[Xiao et~al\mbox{.}(2025)]%
        {xiao2025fine_grained_feedback}
\bibfield{author}{\bibinfo{person}{Wenyi Xiao}, \bibinfo{person}{Ziwei Huang},
  \bibinfo{person}{Leilei Gan}, \bibinfo{person}{Wanggui He},
  \bibinfo{person}{Haoyuan Li}, \bibinfo{person}{Zhelun Yu},
  \bibinfo{person}{Fangxun Shu}, \bibinfo{person}{Hao Jiang}, {and}
  \bibinfo{person}{Linchao Zhu}.} \bibinfo{year}{2025}\natexlab{}.
\newblock \showarticletitle{Detecting and mitigating hallucination in large
  vision language models via fine-grained ai feedback}. In
  \bibinfo{booktitle}{\emph{Proceedings of the AAAI Conference on Artificial
  Intelligence}}, Vol.~\bibinfo{volume}{39}. \bibinfo{pages}{25543--25551}.
\newblock


\bibitem[Yang et~al\mbox{.}(2025)]%
        {yang2025nullu}
\bibfield{author}{\bibinfo{person}{Le Yang}, \bibinfo{person}{Ziwei Zheng},
  \bibinfo{person}{Boxu Chen}, \bibinfo{person}{Zhengyu Zhao},
  \bibinfo{person}{Chenhao Lin}, {and} \bibinfo{person}{Chao Shen}.}
  \bibinfo{year}{2025}\natexlab{}.
\newblock \showarticletitle{Nullu: Mitigating object hallucinations in large
  vision-language models via halluspace projection}. In
  \bibinfo{booktitle}{\emph{Proceedings of the Computer Vision and Pattern
  Recognition Conference}}. \bibinfo{pages}{14635--14645}.
\newblock


\bibitem[Yin et~al\mbox{.}(2024)]%
        {yin2024woodpecker}
\bibfield{author}{\bibinfo{person}{Shukang Yin}, \bibinfo{person}{Chaoyou Fu},
  \bibinfo{person}{Sirui Zhao}, \bibinfo{person}{Tong Xu}, \bibinfo{person}{Hao
  Wang}, \bibinfo{person}{Dianbo Sui}, \bibinfo{person}{Yunhang Shen},
  \bibinfo{person}{Ke Li}, \bibinfo{person}{Xing Sun}, {and}
  \bibinfo{person}{Enhong Chen}.} \bibinfo{year}{2024}\natexlab{}.
\newblock \showarticletitle{Woodpecker: Hallucination correction for multimodal
  large language models}.
\newblock \bibinfo{journal}{\emph{Science China Information Sciences}}
  \bibinfo{volume}{67}, \bibinfo{number}{12} (\bibinfo{year}{2024}),
  \bibinfo{pages}{220105}.
\newblock


\bibitem[Yu et~al\mbox{.}(2024)]%
        {yu2024rlhf}
\bibfield{author}{\bibinfo{person}{Tianyu Yu}, \bibinfo{person}{Yuan Yao},
  \bibinfo{person}{Haoye Zhang}, \bibinfo{person}{Taiwen He},
  \bibinfo{person}{Yifeng Han}, \bibinfo{person}{Ganqu Cui},
  \bibinfo{person}{Jinyi Hu}, \bibinfo{person}{Zhiyuan Liu},
  \bibinfo{person}{Hai-Tao Zheng}, \bibinfo{person}{Maosong Sun},
  {et~al\mbox{.}}} \bibinfo{year}{2024}\natexlab{}.
\newblock \showarticletitle{Rlhf-v: Towards trustworthy mllms via behavior
  alignment from fine-grained correctional human feedback}. In
  \bibinfo{booktitle}{\emph{Proceedings of the IEEE/CVF Conference on Computer
  Vision and Pattern Recognition}}. \bibinfo{pages}{13807--13816}.
\newblock


\bibitem[Zhai et~al\mbox{.}(2023)]%
        {zhai2023halle}
\bibfield{author}{\bibinfo{person}{Bohan Zhai}, \bibinfo{person}{Shijia Yang},
  \bibinfo{person}{Chenfeng Xu}, \bibinfo{person}{Sheng Shen},
  \bibinfo{person}{Kurt Keutzer}, \bibinfo{person}{Chunyuan Li}, {and}
  \bibinfo{person}{Manling Li}.} \bibinfo{year}{2023}\natexlab{}.
\newblock \showarticletitle{HallE-Control: controlling object hallucination in
  large multimodal models}.
\newblock \bibinfo{journal}{\emph{arXiv preprint arXiv:2310.01779}}
  (\bibinfo{year}{2023}).
\newblock


\bibitem[Zhang et~al\mbox{.}(2025a)]%
        {zhang2025openmmreasonerpushingfrontiersmultimodal}
\bibfield{author}{\bibinfo{person}{Kaichen Zhang}, \bibinfo{person}{Keming Wu},
  \bibinfo{person}{Zuhao Yang}, \bibinfo{person}{Bo Li},
  \bibinfo{person}{Kairui Hu}, \bibinfo{person}{Bin Wang},
  \bibinfo{person}{Ziwei Liu}, \bibinfo{person}{Xingxuan Li}, {and}
  \bibinfo{person}{Lidong Bing}.} \bibinfo{year}{2025}\natexlab{a}.
\newblock \bibinfo{title}{OpenMMReasoner: Pushing the Frontiers for Multimodal
  Reasoning with an Open and General Recipe}.
\newblock
\showeprint[arxiv]{2511.16334}~[cs.AI]
\urldef\tempurl%
\url{https://arxiv.org/abs/2511.16334}
\showURL{%
\tempurl}


\bibitem[Zhang et~al\mbox{.}(2024)]%
        {zhang2024vl}
\bibfield{author}{\bibinfo{person}{Ruiyang Zhang}, \bibinfo{person}{Hu Zhang},
  {and} \bibinfo{person}{Zhedong Zheng}.} \bibinfo{year}{2024}\natexlab{}.
\newblock \showarticletitle{Vl-uncertainty: Detecting hallucination in large
  vision-language model via uncertainty estimation}.
\newblock \bibinfo{journal}{\emph{arXiv preprint arXiv:2411.11919}}
  (\bibinfo{year}{2024}).
\newblock


\bibitem[Zhang et~al\mbox{.}(2025b)]%
        {zhang2025dhcp}
\bibfield{author}{\bibinfo{person}{Yudong Zhang}, \bibinfo{person}{Ruobing
  Xie}, \bibinfo{person}{Xingwu Sun}, \bibinfo{person}{Yiqing Huang},
  \bibinfo{person}{Jiansheng Chen}, \bibinfo{person}{Zhanhui Kang},
  \bibinfo{person}{Di Wang}, {and} \bibinfo{person}{Yu Wang}.}
  \bibinfo{year}{2025}\natexlab{b}.
\newblock \showarticletitle{Dhcp: Detecting hallucinations by cross-modal
  attention pattern in large vision-language models}. In
  \bibinfo{booktitle}{\emph{Proceedings of the 33rd ACM International
  Conference on Multimedia}}. \bibinfo{pages}{3555--3564}.
\newblock


\bibitem[Zhao et~al\mbox{.}(2023)]%
        {zhao2023beyond}
\bibfield{author}{\bibinfo{person}{Zhiyuan Zhao}, \bibinfo{person}{Bin Wang},
  \bibinfo{person}{Linke Ouyang}, \bibinfo{person}{Xiaoyi Dong},
  \bibinfo{person}{Jiaqi Wang}, {and} \bibinfo{person}{Conghui He}.}
  \bibinfo{year}{2023}\natexlab{}.
\newblock \showarticletitle{Beyond hallucinations: Enhancing lvlms through
  hallucination-aware direct preference optimization}.
\newblock \bibinfo{journal}{\emph{arXiv preprint arXiv:2311.16839}}
  (\bibinfo{year}{2023}).
\newblock


\bibitem[Zhu et~al\mbox{.}(2023)]%
        {zhu2023minigpt}
\bibfield{author}{\bibinfo{person}{Deyao Zhu}, \bibinfo{person}{Jun Chen},
  \bibinfo{person}{Xiaoqian Shen}, \bibinfo{person}{Xiang Li}, {and}
  \bibinfo{person}{Mohamed Elhoseiny}.} \bibinfo{year}{2023}\natexlab{}.
\newblock \showarticletitle{Minigpt-4: Enhancing vision-language understanding
  with advanced large language models}.
\newblock \bibinfo{journal}{\emph{arXiv preprint arXiv:2304.10592}}
  (\bibinfo{year}{2023}).
\newblock


\bibitem[Zhu et~al\mbox{.}(2025)]%
        {zhu2025ibd}
\bibfield{author}{\bibinfo{person}{Lanyun Zhu}, \bibinfo{person}{Deyi Ji},
  \bibinfo{person}{Tianrun Chen}, \bibinfo{person}{Peng Xu},
  \bibinfo{person}{Jieping Ye}, {and} \bibinfo{person}{Jun Liu}.}
  \bibinfo{year}{2025}\natexlab{}.
\newblock \showarticletitle{Ibd: Alleviating hallucinations in large
  vision-language models via image-biased decoding}. In
  \bibinfo{booktitle}{\emph{Proceedings of the Computer Vision and Pattern
  Recognition Conference}}. \bibinfo{pages}{1624--1633}.
\newblock


\bibitem[Zollicoffer et~al\mbox{.}(2025)]%
        {zollicoffer2025mtre}
\bibfield{author}{\bibinfo{person}{Geigh Zollicoffer}, \bibinfo{person}{Minh
  Vu}, {and} \bibinfo{person}{Manish Bhattarai}.}
  \bibinfo{year}{2025}\natexlab{}.
\newblock \showarticletitle{MTRE: Multi-Token Reliability Estimation for
  Hallucination Detection in VLMs}.
\newblock \bibinfo{journal}{\emph{arXiv preprint arXiv:2505.11741}}
  (\bibinfo{year}{2025}).
\newblock


\end{thebibliography}

\clearpage
\appendix

\tcbset{before skip=6pt, after skip=6pt}

\section{Dataset Details}
\label{app:dataset}
\subsection{Dataset Composition}
\label{app:dataset-composition}

Table~\ref{tab:app_dataset} summarizes the composition of HalluScope-30K. The dataset is split into a training set of 27,666 samples and a test set of 733 samples. Both sets cover all eight source datasets across five task categories, ensuring that the test set evaluates the full spectrum of hallucination types. The eight source datasets are:
\begin{itemize}[leftmargin=*,nosep]
\item \textbf{MMBench}~\cite{liu2024mmbench}: A multi-discipline visual understanding benchmark covering perception, reasoning, and knowledge across multiple languages.
\item \textbf{MMStar}~\cite{chen2024mmstar}: A curated multimodal benchmark emphasizing vision-critical questions that resist text-only shortcuts.
\item \textbf{MathV360K}~\cite{shi2024math}: A large-scale mathematical visual QA collection spanning 24 sub-domains including geometry, algebra, and statistics.
\item \textbf{Geo170K}~\cite{gao2023g}: A geometry-focused dataset with diagram-based reasoning questions from Geo3K and GeoQA+.
\item \textbf{OCRBench}~\cite{liu2024ocrbench}: An OCR evaluation benchmark covering text recognition, scene text understanding, and document analysis.
\item \textbf{RealWorldQA}~\cite{xai2024grok}: Real-world spatial understanding questions requiring reasoning about object positions, orientations, and environments.
\item \textbf{M-HalDetect}~\cite{gunjal2024mhal}: An image captioning and VQA dataset with COCO images, originally designed for multimodal hallucination detection.
\item \textbf{RLHF-V}~\cite{yu2024rlhf}: A visual QA dataset with human preference annotations; we use the chosen answer as ground truth.
\end{itemize}

\begin{table}[h]
\centering
\caption{Composition of HalluScope-30K by source dataset and task category.}
\label{tab:app_dataset}
\setlength{\tabcolsep}{5pt}
\renewcommand{\arraystretch}{1.15}
\small
\begin{tabular}{l|l|rrr}
\toprule
\textbf{Task Category} & \textbf{Source Dataset} & \textbf{Train} & \textbf{Test} & \textbf{Total} \\
\midrule
\multirow{2}{*}{General Perception}
 & MMBench & 5,026 & 113 & 5,139 \\
 & MMStar & 648 & 58 & 706 \\
\midrule
\multirow{2}{*}{\shortstack[l]{Mathematical\\Reasoning}}
 & MathV360K & 6,064 & 123 & 6,187 \\
 & Geo170K & 3,898 & 123 & 4,021 \\
\midrule
OCR & OCRBench & 792 & 44 & 836 \\
\midrule
Spatial Reasoning & RealWorldQA & 496 & 42 & 538 \\
\midrule
\multirow{2}{*}{\shortstack[l]{Hallucination\\Detection}}
 & M-HalDetect & 6,645 & 114 & 6,759 \\
 & RLHF-V & 4,097 & 116 & 4,213 \\
\midrule
\multicolumn{2}{l|}{\textbf{Total}} & \textbf{27,666} & \textbf{733} & \textbf{28,399} \\
\bottomrule
\end{tabular}
\end{table}

\subsection{Data Preprocessing}
\label{app:preprocessing}

Each of the source datasets listed in Table~\ref{tab:app_dataset} undergoes a standardized preprocessing pipeline before entering the hallucination label generation stage. The preprocessing consists of the following steps:

\begin{enumerate}
\item \textbf{Format unification.} Raw datasets differ in schema and annotation style. We convert all samples into a unified format with fields: image, question, answer, question type (MCQ or open-ended), and source metadata.
\item \textbf{Subset sampling.} For large-scale datasets (MathV360K and Geo170K), we use randomly sampled subsets to maintain a balanced composition across task categories.
\item \textbf{Deduplication.} We construct a deduplication key from the concatenation of question, choices, and answer, and remove duplicate entries keeping the first occurrence.
\end{enumerate}

After preprocessing, all samples enter the hallucination label generation pipeline. The final HalluScope-30K is obtained after multi-stage quality control (Appendix~\ref{app:quality-control}).

\subsection{Injection vs.\ Annotation Ratio}
\label{app:injection-annotation}

As described in the main text, HalluScope employs two complementary label generation strategies based on model response correctness: \textit{hallucination injection} for correctly answered samples and \textit{hallucination annotation} for incorrectly answered ones. Table~\ref{tab:app_injection_ratio} reports the ratio of samples processed through each pathway.

\begin{table}[h]
\centering
\caption{Ratio of hallucination injection (correct answers) vs.\ annotation (incorrect answers) by task category.}
\label{tab:app_injection_ratio}
\setlength{\tabcolsep}{4pt}
\renewcommand{\arraystretch}{1.15}
\begin{tabular}{lccc}
\toprule
\textbf{Task Category} & \textbf{Injection} & \textbf{Annotation} & \textbf{Inject. \%} \\
\midrule
General Perception & 4,577 & 1,097 & 80.7\% \\
Math.\ Reasoning & 7,283 & 2,679 & 73.1\% \\
OCR & 691 & 101 & 87.2\% \\
Spatial Reasoning & 326 & 170 & 65.7\% \\
Halluc.\ Detection & 0 & 10,742 & 0.0\% \\
\midrule
\textbf{Total} & \textbf{12,877} & \textbf{14,789} & \textbf{46.5\%} \\
\bottomrule
\end{tabular}
\end{table}

The ratio varies substantially across task categories. OCR and general perception datasets exhibit high injection rates of 87.2\% and 80.7\%, respectively, because the base model Qwen3-VL-8B-Instruct answers most of these questions correctly. Mathematical reasoning has a moderate injection rate of 73.1\% due to the model's lower accuracy on geometry and math problems. Hallucination detection datasets, namely M-HalDetect and RLHF-V, are entirely processed through the annotation pathway, as these datasets provide pre-existing model responses that already contain hallucinations.

\subsection{Hallucination Type Distribution}
\label{app:type-distribution}

\begin{figure*}[h]
  \centering
  \includegraphics[width=\textwidth]{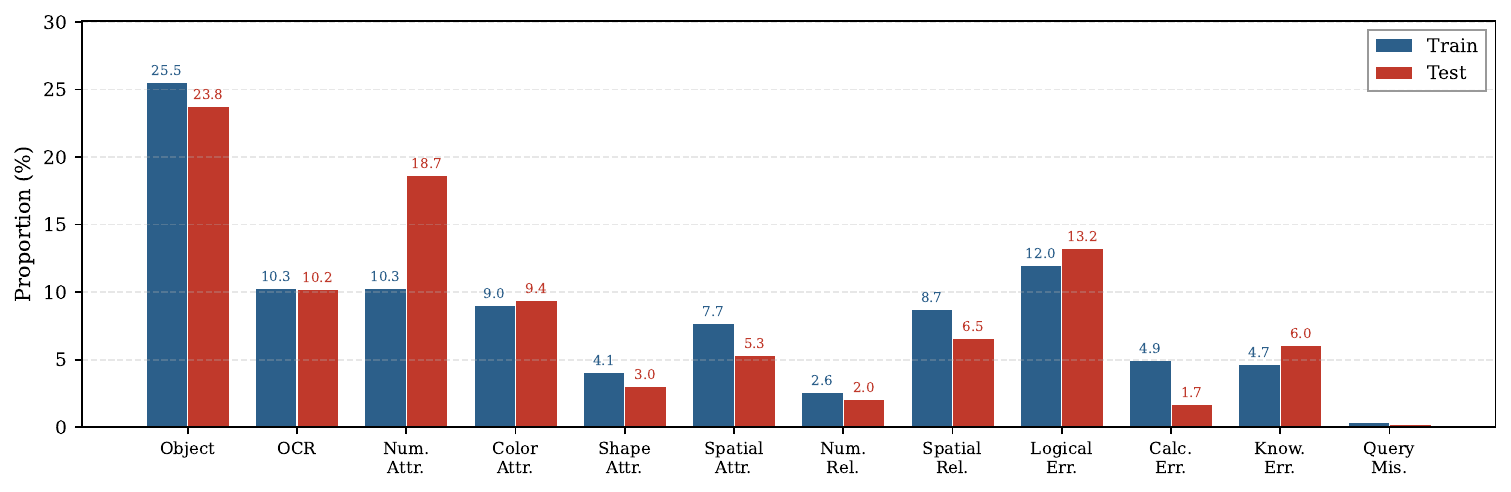}
  \caption{Proportion of each hallucination type in the training set (27,666 samples, 141,156 spans) and the test set (733 samples, 3,375 spans).}
  \label{fig:app_type_dist}
  \Description{Bar chart comparing the proportion of 12 hallucination types between the training and test sets of HalluScope-30K.}
\end{figure*}

Figure~\ref{fig:app_type_dist} illustrates the distribution of hallucination types in the training and test sets. Each sample may contain multiple hallucination spans of different types, so the span counts exceed the total number of samples.

The distribution exhibits notable class imbalance. \textit{Object} is the most frequent type, accounting for 25.5\% of all training spans, while \textit{Query\_Misunderstanding} is extremely underrepresented with only 486 training spans (0.34\%) and 7 test spans. This imbalance directly explains the model's low performance on these rare types (e.g., Query\_Misunderstanding). On average, each sample contains 5.1 hallucination spans in the training set and 4.6 in the test set.

\begin{figure}[h]
  \centering
  \includegraphics[width=0.65\columnwidth]{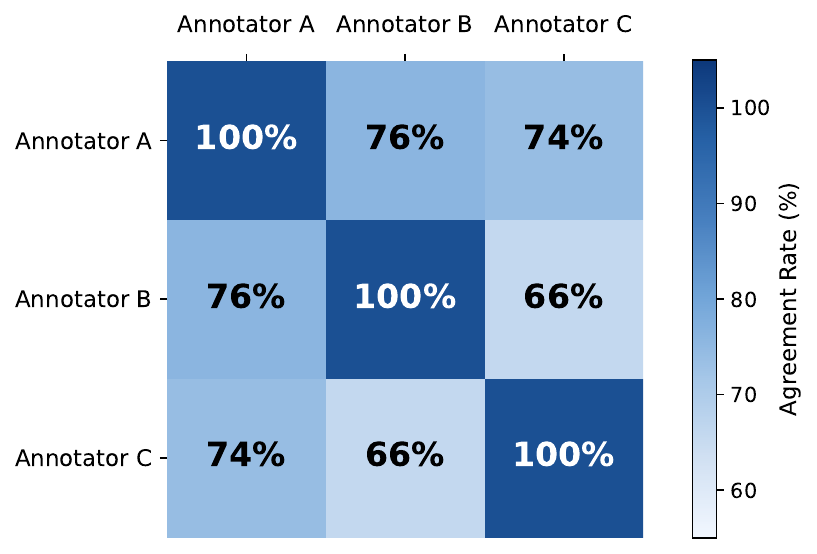}
  \caption{Pairwise agreement rates among three annotators on 100 randomly sampled test samples.}
  \label{fig:agreement_matrix}
  \Description{A 3x3 heatmap showing pairwise inter-annotator agreement rates: A--B 76.0\%, A--C 74.0\%, B--C 66.0\%.}
\end{figure}

\subsection{Benchmark Construction}
\label{app:benchmark-construction}

The hallucination classification benchmark consists of 733 samples drawn from the same data generation pipeline as the training split, with strict deduplication to ensure no overlap between the two splits. The per-subset composition is shown in Table~\ref{tab:app_dataset}, covering all task categories. Each test sample undergoes the same quality control process as the training data, including text quality assessment and hallucination consistency verification.

To assess the annotation quality of the benchmark, three independent annotators each evaluated 100 randomly sampled test instances and judged whether the hallucination labels (span boundaries and type assignments) were correct. Note that this annotation task requires jointly verifying both the span boundary and the hallucination type from a fine-grained taxonomy of 12 categories, which is substantially more demanding than binary hallucination detection. Annotators~A and~B each reported an accuracy of 80.0\%, while Annotator~C reported 76.0\%. The pairwise agreement rates are 76.0\% (A--B), 74.0\% (A--C), and 66.0\% (B--C), as shown in Figure~\ref{fig:agreement_matrix}. The lower pairwise rates primarily reflect the inherent subjectivity in fine-grained type assignment rather than span detection errors: annotators typically agreed on whether a span is hallucinated but occasionally diverged on the assigned type (e.g., \textit{Numerical\_Attribute} vs.\ \textit{Calculation\_Error}). Under majority voting (at least two annotators agree), the accuracy reaches \textbf{86.0\%}, confirming the high quality of the benchmark annotations.

\section{Hallucination Taxonomy}
\label{app:taxonomy}

\begin{table*}[t]
\centering
\definecolor{TaxHeader}{RGB}{214,226,240}
\definecolor{ObjMask}{HTML}{FFCDD2}
\definecolor{OCRMask}{HTML}{F8BBD0}
\definecolor{ColorMask}{HTML}{E1BEE7}
\definecolor{SpatialMask}{HTML}{D1C4E9}
\definecolor{NumericalMask}{HTML}{C5CAE9}
\definecolor{ShapeMask}{HTML}{BBDEFB}
\definecolor{LogicalMask}{HTML}{B2DFDB}
\definecolor{CalculationMask}{HTML}{C8E6C9}
\definecolor{KnowledgeMask}{HTML}{DCEDC8}
\definecolor{QueryMask}{HTML}{FFF9C4}
\caption{Hallucination types and definitions used for hallucination diagnosis, adopted from the taxonomy of MHALO~\cite{cai2025mhalo}. Types above the mid-rule are perception-level; types below are reasoning-level.}
\label{tab:hallucination_taxonomy}
\renewcommand{\arraystretch}{1.18}
\resizebox{\textwidth}{!}{%
\begin{tabular}{lp{13.8cm}}
\toprule
\rowcolor{TaxHeader}
\textbf{Type} & \textbf{Definition} \\
\midrule
\colorbox{ObjMask}{Object} & Errors in recognizing an object's presence, absence, or category. \\
\colorbox{OCRMask}{OCR} & Failure to correctly read visible text, numbers, or symbols. \\
\colorbox{NumericalMask}{Numerical Attribute} & Errors in identifying the absolute value of a single object or element. \\
\colorbox{ColorMask}{Color Attribute} & Misperception of the color of a correctly identified object or region. \\
\colorbox{ShapeMask}{Shape Attribute} & Errors in describing an object's geometric or structural shape. \\
\colorbox{SpatialMask}{Spatial Attribute} & Errors in identifying a single object's position, orientation, or distance. \\
\midrule
\colorbox{NumericalMask}{Numerical Relation} & Errors in reasoning about quantitative relations among objects or values. \\
\colorbox{SpatialMask}{Spatial Relation} & Errors in reasoning about spatial relations among multiple objects. \\
\colorbox{LogicalMask}{Logical Error} & Wrong conclusions caused by flawed logical inference. \\
\colorbox{CalculationMask}{Calculation Error} & Mistakes in arithmetic or mathematical computation. \\
\colorbox{KnowledgeMask}{Knowledge Error} & Incorrect use of external knowledge, common sense, or domain rules. \\
\colorbox{QueryMask}{Query Misunderstanding} & Responses caused by misunderstanding the query intent or constraints. \\
\bottomrule
\end{tabular}%
}
\end{table*}

We adopt the hallucination taxonomy proposed by MHALO~\cite{cai2025mhalo}, which organizes 12 fine-grained hallucination types into two high-level categories: \textit{perception-level} and \textit{reasoning-level}. Table~\ref{tab:hallucination_taxonomy} summarizes all types with their definitions.

\noindent\textbf{Perception-level hallucinations} arise from errors in visual understanding or information extraction, and can be directly verified by examining the image. They are organized into two sub-groups:

\begin{itemize}[leftmargin=*,nosep]
\item \textbf{Content errors}: \textit{Object}, which refers to misidentifying an object's presence, absence, or category, and \textit{OCR}, which refers to misreading visible text, numbers, or symbols.
\item \textbf{Attribute errors}: \textit{Numerical\_Attribute} for wrong absolute count or measurement of a single entity, \textit{Color\_Attribute} for wrong color of a correctly identified object, \textit{Shape\_Attribute} for wrong geometric or structural form, and \textit{Spatial\_Attribute} for wrong position, orientation, or distance of a single object.
\end{itemize}

\noindent\textbf{Reasoning-level hallucinations} originate from incorrect inference, computation, or knowledge application, requiring cognitive processing beyond direct visual observation. They are organized into two sub-groups:

\begin{itemize}[leftmargin=*,nosep]
\item \textbf{Relation errors}: \textit{Numerical\_Relation} for wrong quantitative comparison between multiple entities, and \textit{Spatial\_Relation} for wrong spatial arrangement between multiple objects. Although relation errors may involve visual content, they require comparative reasoning across multiple objects rather than direct perception of a single element.
\item \textbf{Reasoning errors}: \textit{Logical\_Error} for flawed reasoning chains or invalid inferences, \textit{Calculation\_Error} for mistakes in arithmetic or mathematical computation, \textit{Knowledge\_Error} for incorrect application of external knowledge or domain-specific facts, and \textit{Query\_Misunderstanding} for responding to a different question than what was asked.
\end{itemize}

\section{Quality Control Details}
\label{app:quality-control}

\subsection{Label Normalization}
\label{app:label-normalization}

The label normalization stage maps non-standard hallucination type labels produced by the LLM annotator to the standardized set of 12 types.
The pipeline applies over 80 rule-based mappings, including
case normalization (e.g., ``object'' $\to$ ``Object''),
plural-to-singular conversion (e.g., ``Calculation\_Errors'' $\to$ ``Calculation\_Error''),
format unification for space-separated and CamelCase variants,
and semantic synonym mapping
(e.g., ``Entity'' $\to$ ``Object'',
``Factual\_Error'' $\to$ ``Knowledge\_Error'').
If no rule matches, fuzzy matching is attempted by comparing labels after stripping underscores, spaces, and hyphens.
Samples containing unmappable labels (e.g., ``Unknown'', ``None'') are discarded entirely.

\subsection{Text Quality Assessment}
\label{app:text-quality}

We use LLM-as-Judge to assess the text quality of each generated sample. The judge evaluates three dimensions on a 1--5 scale:

\begin{itemize}[leftmargin=*,nosep]
\item \textbf{Clarity} (1--5): Whether the response is easy to understand, with unambiguous language and clear expression of ideas.
\item \textbf{Fluency} (1--5): Whether the response reads naturally, with proper grammar, word choice, and sentence flow.
\item \textbf{Coherence} (1--5): Whether the response maintains logical consistency throughout, with smooth transitions between ideas and no contradictions.
\end{itemize}

A sample passes quality assessment only if it scores $\geq 4$ on all three dimensions. Samples scoring below the threshold on any dimension are removed from the dataset.

\subsection{Hallucination Consistency Verification}
\label{app:consistency-verification}

After label normalization and text quality assessment, we use LLM-as-Judge to verify the consistency of the generated hallucination labels. Specifically, for each sample, the judge model is provided with the original image, question, ground-truth answer, and the hallucination-annotated response, and is asked to evaluate two aspects: (1) whether each tagged hallucination is semantically plausible as a realistic MLLM error, and (2) whether the surrounding text remains coherent after hallucination injection. In addition, rule-based checks are applied to ensure format correctness (proper XML tag closure, no nesting) and type validity (each type attribute belongs to the predefined 12-type set). Samples that fail any of these checks are removed from the final dataset.

\section{Implementation Details}
\label{app:implementation}

\subsection{Training Hyperparameters}
\label{app:hyperparams}

All experiments are conducted on four servers, including two servers equipped with 2 NVIDIA H20Z GPUs (144\,GB) and two servers equipped with 2 NVIDIA B20Z GPUs (180\,GB). We use Qwen3-VL-4B-Instruct and Qwen3-VL-8B-Instruct as the base models.

\noindent\textbf{SFT stage.} We apply LoRA fine-tuning with rank $r=8$ and alpha $\alpha=16$, targeting all linear layers with the vision tower frozen. Training runs for 1 epoch with a learning rate of $5\times10^{-5}$, a batch size of 8, and a maximum sequence length of 4096. After SFT, the LoRA adapters are merged into the base model.

\noindent\textbf{RL stage.} We apply GRPO with a group size $K=5$ and full-parameter training (no LoRA) with the vision tower unfrozen. The learning rate is $1\times10^{-6}$, global batch size is 32, and training runs for 250 steps. The KL coefficient is 0.01 and the reward weight $\lambda$ is set to 0.3. The maximum prompt length is 4096 and maximum response length is 2048. During training rollout, temperature is 1.0 and top-$p$ is 1.0; during validation, temperature is 0.6 and top-$p$ is 0.95. We use FSDP with CPU offloading and vLLM for rollout generation. The RL framework is based on EasyR1 (built on veRL).

\begin{figure*}[t]
  \centering
  \includegraphics[width=\textwidth]{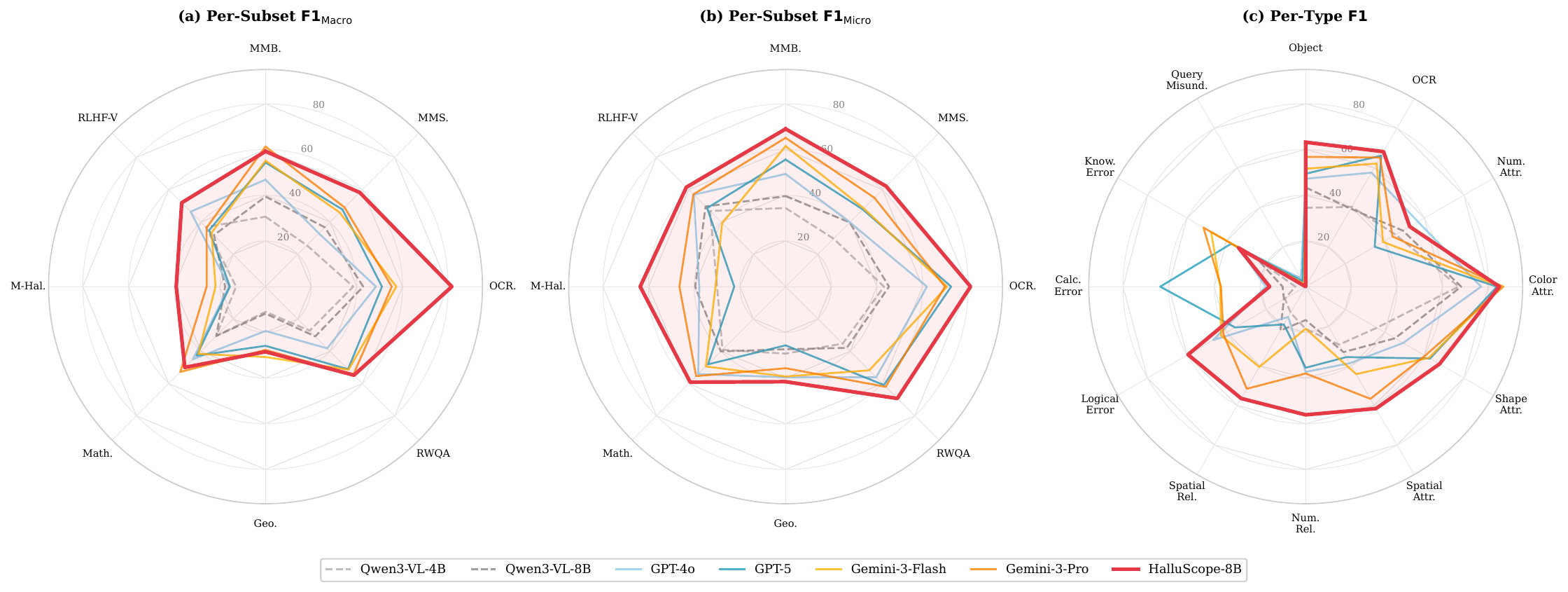}
  \caption{Per-subset and per-type classification results on our hallucination classification benchmark. (a) Per-subset $\mathbf{F1}_{\mathrm{Macro}}$. (b) Per-subset $\mathbf{F1}_{\mathrm{Micro}}$. (c) Per-type F1 across 12 hallucination types.}
  \label{fig:three_radar}
  \Description{Three radar charts showing per-subset Macro F1, per-subset Micro F1, and per-type F1 scores across seven models. HalluScope-8B (red) achieves the largest coverage area in all three charts.}
\end{figure*}

\begin{table}[t]
\centering
\caption{Effect of $\lambda$ on detection and classification performance. Best results per metric are \textbf{bolded}.}
\label{tab:lambda-ablation}
\renewcommand{\arraystretch}{1.1}

\begin{tabular*}{\columnwidth}{@{\extracolsep{\fill}}cccccc@{}}
\toprule
\multirow{2}{*}{$\lambda$}
& \multicolumn{2}{c}{MHALO}
& \multicolumn{3}{c}{Classification} \\
\cmidrule(lr){2-3} \cmidrule(lr){4-6}
& Avg $\mathbf{F1}_{\mathrm{M}}$
& Avg $\mathbf{F1}_{\mathrm{IoU}}$
& $\mathbf{F1}_{\mathrm{M}}$
& $\mathbf{F1}_{\mathrm{Macro}}$
& $\mathbf{F1}_{\mathrm{Micro}}$ \\
\midrule
0.1 & 61.97 & 56.65 & \textbf{72.48} & 49.74 & 58.54 \\
0.3 & 64.03 & 57.57 & 72.20 & 51.66 & \textbf{59.60} \\
0.5 & 62.88 & 56.98 & 71.14 & \textbf{52.53} & 59.55 \\
0.7 & \textbf{64.92} & \textbf{59.10} & 70.70 & 51.87 & 58.88 \\
\bottomrule
\end{tabular*}
\end{table}

\subsection{Mitigation Experiment Setup}
\label{app:mitigation-setup}

We provide additional details on the mitigation experiment. The initial accuracy before feedback is 68.76\% (504/733) for Qwen3-VL-8B-Instruct and 33.02\% (242/733) for LLaVA-1.5-7B. All models use greedy decoding (temperature\,=\,0) with a maximum generation length of 2048 tokens. For free-form questions, correctness is determined by Gemini-3-Pro judge through semantic equivalence; for MCQs, the option letter is extracted and compared directly. Feedback prompt templates for all four modes are provided in Appendix~\ref{app:prompt-feedback}.

\section{Evaluation Metrics}
\label{app:metrics}

\noindent\textbf{$\mathbf{F1}_{\mathrm{Macro}}$:}
\begin{equation}
\mathbf{F1}_{\mathrm{Macro}}
=
\frac{1}{C}
\sum_{k=1}^{C}
\frac{2TP_k}{2TP_k + FP_k + FN_k}.
\end{equation}

This metric is insensitive to class frequency and better reflects balanced performance across hallucination types.

\noindent\textbf{$\mathbf{F1}_{\mathrm{Micro}}$:}
\begin{equation}
\mathbf{F1}_{\mathrm{Micro}}
=
\frac{2\sum_{k=1}^{C} TP_k}
{2\sum_{k=1}^{C} TP_k + \sum_{k=1}^{C} FP_k + \sum_{k=1}^{C} FN_k }.
\end{equation}

$\mathbf{F1}_{\mathrm{Micro}}$ is computed from globally aggregated statistics and is more influenced by frequent classes.

\begin{table}[h]
\centering
\caption{Multiplicative vs. additive detection reward composition. Best results per metric are \textbf{bolded}.}
\label{tab:reward-composition}
\renewcommand{\arraystretch}{1.1}
\begin{tabular}{lcccccc}
\toprule
\multirow{2}{*}{Composition} & \multicolumn{2}{c}{MHALO} & \multicolumn{3}{c}{Classification} \\
\cmidrule(lr){2-3} \cmidrule(lr){4-6}
& Avg $\mathbf{F1}_{\mathrm{M}}$ & Avg $\mathbf{F1}_{\mathrm{IoU}}$ & $\mathbf{F1}_{\mathrm{M}}$ & $\mathbf{F1}_{\mathrm{Macro}}$ & $\mathbf{F1}_{\mathrm{Micro}}$ \\
\midrule
Multiplicative & \textbf{64.03} & \textbf{57.57} & \textbf{72.20} & 51.66 & \textbf{59.60} \\
Additive & 62.70 & 57.36 & 71.13 & \textbf{51.99} & 59.46 \\
\bottomrule
\end{tabular}
\end{table}

\section{Additional Experimental Results}
\label{app:results}

Figure~\ref{fig:three_radar} presents the per-subset and per-type classification results on our hallucination classification benchmark.

\noindent\textbf{Per-subset analysis.} HalluScope-8B achieves the highest $\mathbf{F1}_{\mathrm{Micro}}$ on all eight subsets. The most pronounced advantage appears on OCRBench ($\mathbf{F1}_{\mathrm{Macro}}$=81.6 vs.\ 57.3 for Gemini-3-Flash) and M-HalDetect ($\mathbf{F1}_{\mathrm{Micro}}$=63.7 vs.\ 46.5 for Gemini-3-Pro), indicating that our model particularly excels at OCR-related and open-ended hallucination classification. Geo170K remains the most challenging subset for all models ($\mathbf{F1}_{\mathrm{Micro}} < 42$), reflecting the difficulty of fine-grained type classification on geometry problems.

\noindent\textbf{Per-type analysis.} HalluScope-8B achieves the best F1 on 7 of 12 hallucination types. The largest gains appear on reasoning-level types: \textit{Numerical\_Relation} (56.1 vs.\ 38.0 for Gemini-3-Pro), \textit{Logical\_Error} (59.5 vs.\ 41.6), and \textit{Spatial\_Relation} (56.5 vs.\ 51.6). A notable weakness is \textit{Calculation\_Error} (F1=15.8), where GPT-5 reaches 63.7, suggesting that numerical computation verification benefits more from the stronger reasoning capabilities of larger models. \textit{Query\_Misunderstanding} yields F1=0.0 for both HalluScope-8B and Gemini-3-Pro, consistent with its extreme scarcity in the dataset (only 7 test spans, see Appendix~\ref{app:type-distribution}).

\subsection{Explanation Quality Evaluation}
\label{app:explanation-quality}

Beyond detection and classification, HalluScope also generates natural language explanations that describe why a hallucination occurs and how it may be corrected. To directly assess the quality of these explanations, we randomly sample 50 diagnosed cases from our classification benchmark and invite three independent annotators to rate each explanation on a 1--5 scale along three dimensions: \textbf{Causal Accuracy (CA)}, whether the explanation correctly identifies the cause of the hallucination; \textbf{Correction Usefulness (CU)}, whether it provides actionable correction guidance; and \textbf{Visual Grounding (VG)}, whether the reasoning is grounded in image evidence.

As shown in Table~\ref{tab:explanation_quality}, the explanations receive an average score of 3.89, with particularly high Visual Grounding, indicating that the diagnostic reasoning is well grounded in the visual input. The inter-annotator agreement (pairwise agreement within one point) reaches 70.7\% for CA, 82.0\% for CU, and 86.0\% for VG, indicating good consistency across all three dimensions. These results directly validate the quality of HalluScope's diagnostic explanations.

\begin{table}[t]
\centering
\caption{Human evaluation of explanation quality on 50 randomly sampled diagnosed cases. CA: Causal Accuracy; CU: Correction Usefulness; VG: Visual Grounding.}
\label{tab:explanation_quality}
\renewcommand{\arraystretch}{1.12}
\begin{tabular*}{\columnwidth}{@{\extracolsep{\fill}}lcccc@{}}\toprule
\textbf{Annotator} & CA$\uparrow$ & CU$\uparrow$ & VG$\uparrow$ & Avg$\uparrow$ \\
\midrule
Annotator \#1 & 3.90 & 3.62 & 4.42 & 3.98 \\
Annotator \#2 & 3.40 & 3.28 & 4.46 & 3.71 \\
Annotator \#3 & 3.72 & 3.24 & 4.96 & 3.97 \\
\midrule
\textbf{Average} & \textbf{3.67} & \textbf{3.38} & \textbf{4.61} & \textbf{3.89} \\
\bottomrule
\end{tabular*}
\end{table}

\section{Additional Ablation Studies}
\label{app:ablation}

\subsection{Reward Weight Sensitivity ($\lambda$)}
\label{app:lambda-ablation}

As described in the main text, the RL stage optimizes a multi-granular joint reward that combines a span-level detection reward $r_{\mathrm{det}}(y)$ with a type-level classification reward $r_{\mathrm{cls}}(y)$. The hyperparameter $\lambda$ controls the relative importance of these two objectives. A higher $\lambda$ allocates more training signal toward correct type classification, while a lower $\lambda$ prioritizes accurate span detection. The multi-granular joint reward is formulated as:
\begin{equation}
    R(y)=\mathbb{I}_{\mathrm{format}}\left((1-\lambda)\, r_{\mathrm{det}}(y)+\lambda\, r_{\mathrm{cls}}(y)\right)
\end{equation}
We train models with $\lambda \in \{0.1, 0.3, 0.5, 0.7\}$ and report results on both the MHALO benchmark and our hallucination classification benchmark in Table~\ref{tab:lambda-ablation}.

Increasing $\lambda$ allocates more training signal toward type classification. As shown in Table~\ref{tab:lambda-ablation}, $\lambda=0.7$ achieves the highest MHALO detection scores with Avg $\mathbf{F1}_{\mathrm{M}}$=64.92 and Avg $\mathbf{F1}_{\mathrm{IoU}}$=59.10, likely because the classification reward $r_{\mathrm{cls}}$ also implicitly encourages accurate span detection through type-matched IoU. However, it degrades performance on our hallucination classification benchmark to $\mathbf{F1}_{\mathrm{M}}$=70.70. Conversely, $\lambda=0.1$ maximizes $\mathbf{F1}_{\mathrm{M}}$ at 72.48 but yields the weakest MHALO detection at 61.97 and the lowest $\mathbf{F1}_{\mathrm{Macro}}$ of 49.74. We select $\lambda=0.3$ as the default because it achieves competitive MHALO detection at 64.03 while maintaining strong classification performance across all three metrics: $\mathbf{F1}_{\mathrm{M}}$=72.20, $\mathbf{F1}_{\mathrm{Macro}}$=51.66, and $\mathbf{F1}_{\mathrm{Micro}}$=59.60, providing the best overall balance.

\subsection{Detection Reward Composition: Multiplicative vs. Additive}
\label{app:reward-composition}

We compare two composition modes for the detection reward: the multiplicative form $r_{\mathrm{det}} = r_{\mathrm{content}} \times r_{\mathrm{count}}$ and an additive form $r_{\mathrm{det}} = 0.5 \cdot r_{\mathrm{content}} + 0.5 \cdot r_{\mathrm{count}}$. Table~\ref{tab:reward-composition} summarizes the results.

The multiplicative form outperforms the additive variant on most metrics, with a notable advantage on MHALO detection where $\mathbf{F1}_{\mathrm{M}}$ reaches 64.03 vs.\ 62.70 for the additive form. This is expected because the multiplicative composition requires both content matching and count accuracy to be high simultaneously, providing a stronger training signal that discourages partial matches.

\onecolumn
\section{Prompt Templates}
\label{app:prompts}

\tcbset{before skip=6pt, after skip=6pt}

\subsection{Data Generation Prompts}
\label{app:prompt-datagen}

Our data generation pipeline uses four prompt-driven stages: Stage~1 generates model answers and judges correctness, Stage~2 applies hallucination label generation through injection or annotation, Stage~3 assesses linguistic quality, and Stage~4 verifies hallucination consistency. The full prompts are listed below.

\begin{tcolorbox}[breakable,colback=sysfill, colframe=black!30, boxrule=0.3pt,
  sharp corners, left=0.025\textwidth, right=0.025\textwidth, top=1pt, bottom=1pt,
  fonttitle=\footnotesize\bfseries,
  title={Stage 1a: Model Answer Generation},
  toptitle=0.5pt, bottomtitle=0.5pt,
  coltitle=black, colbacktitle=systitle]
\small\sffamily
You are an expert visual question answering assistant.

\textbf{Question:} \texttt{\{question\}}

\textbf{Instructions:} Analyze the image carefully and think step by step.\\
1. Generate a step-by-step reasoning process that naturally leads to the correct answer\\
2. Show your complete thought process and analysis\\
3. Make sure your reasoning is logical, coherent, and educational\\
4. Finally, conclude with the provided correct answer

\textbf{Important:}\\
-- Be thorough in your reasoning and analysis\\
-- Consider all relevant information from the question\\
-- Your reasoning should naturally lead to the answer\\
-- Make the reasoning process natural and instructive

Please provide your reasoning and answer.
\end{tcolorbox}

\begin{tcolorbox}[breakable,colback=sysfill, colframe=black!30, boxrule=0.3pt,
  sharp corners, left=0.025\textwidth, right=0.025\textwidth, top=1pt, bottom=1pt,
  fonttitle=\footnotesize\bfseries,
  title={Stage 1b: Answer Correctness Judge},
  toptitle=0.5pt, bottomtitle=0.5pt,
  coltitle=black, colbacktitle=systitle]
\small\sffamily
Determine if the model's answer is semantically correct compared to the ground truth.\\
-- For numerical answers: exact match required\\
-- For descriptive answers: semantic equivalence is sufficient

\textbf{Question:} \texttt{\{question\}}\\
\textbf{Ground Truth:} \texttt{\{ground\_truth\}}\\
\textbf{Model Answer:} \texttt{\{model\_answer\}}

\textbf{Instructions:} Determine if the Model Answer is consistent with the Ground Truth Answer. Consider:\\
1. The core meaning/result should be the same\\
2. Numerical answers should match exactly\\
3. For multiple choice, the selected option should match

Respond with ONLY one of the following: ``\textbf{CORRECT}'' or ``\textbf{INCORRECT}''
\end{tcolorbox}

\begin{tcolorbox}[breakable,colback=userfill, colframe=black!30, boxrule=0.3pt,
  sharp corners, left=0.025\textwidth, right=0.025\textwidth, top=1pt, bottom=1pt,
  fonttitle=\footnotesize\bfseries,
  title={Stage 2a: Hallucination Injection (for correct samples)},
  toptitle=0.5pt, bottomtitle=0.5pt,
  coltitle=black, colbacktitle=usertitle]
\small\sffamily
You are an expert at injecting hallucinations into visual question answering data.

\textbf{Task:} Inject 3--5 different types of hallucinations into the given correct model answer to create a plausible but incorrect response. Each hallucination tag should cover a natural semantic unit --- use short phrases (1--3 words) when possible, but longer spans are acceptable when the error is a coherent phrase.

\textbf{Hallucination Types:} \textit{[12 types with definitions and boundary rules]}

\textbf{Dataset Category:} \texttt{\{dataset\_category\}}\\
\textbf{Priority Types (use 2--3):} \texttt{\{primary\_types\}}\\
\textbf{Secondary Types (use 1--2):} \texttt{\{secondary\_types\}}

\textbf{Injection Strategies:} Apply one or more of the following strategies to inject hallucinations naturally:\\
1.~\textbf{Entity Replacement}: Replace correct entities with similar but incorrect ones.\\
2.~\textbf{Object Attribute Modification}: Change attributes while keeping the object.\\
3.~\textbf{Object Relationship Modification}: Change relationships between multiple objects.\\
4.~\textbf{Numerical Perturbation}: Modify numerical values.\\
5.~\textbf{Precision-to-Vagueness Conversion}: Replace precise info with misleading vague expressions.\\
6.~\textbf{Knowledge Substitution}: Replace correct domain knowledge with plausible but incorrect info.\\
7.~\textbf{Reasoning Chain Disruption}: Modify or break the logical flow of reasoning.\\
8.~\textbf{OCR-Specific Errors}: Introduce text/number recognition errors.\\
9.~\textbf{Spatial Position Perturbation}: Modify spatial descriptions.\\
10.~\textbf{Query Intent Redirection}: Answer a different but related question.\\
11.~\textbf{Comparative Relation Inversion}: Reverse comparison relationships.\\
12.~\textbf{Multi-Step Calculation Error}: Introduce errors at specific steps in multi-step calculations.

\textbf{Input:}\\
-- Question: \texttt{\{question\}}\\
-- Ground Truth Answer: \texttt{\{ground\_truth\}}\\
-- Model Answer (Correct): \texttt{\{model\_answer\}}

\textbf{Requirements:}\\
1.~Inject 3--5 different hallucination types. Total tags can be more than 5.\\
2.~Use format: \texttt{<hallucination type="TYPE">hallucinated content</hallucination>}.\\
3.~Tags must not be nested or overlapping. Adjacent tags with the SAME type MUST be merged. Adjacent tags with DIFFERENT types are allowed.\\
4.~Make hallucinations plausible and natural.\\
5.~Each tag should cover a natural semantic unit --- single word, short phrase (2--3 words), or medium phrase (4--5 words).\\
6.~Maintain text consistency after tag insertion.\\
7.~Diversify hallucination types; do not use a single type throughout.\\
8.~Vary tag lengths; mix single-word and short-phrase tags.\\
9.~Only merge adjacent hallucinated words when they belong to the same type.

\textbf{Output:} The complete modified answer with all hallucination tags.
\end{tcolorbox}

\begin{tcolorbox}[breakable,colback=outfill, colframe=black!30, boxrule=0.3pt,
  sharp corners, left=0.025\textwidth, right=0.025\textwidth, top=1pt, bottom=1pt,
  fonttitle=\footnotesize\bfseries,
  title={Stage 2b: Hallucination Annotation (for incorrect samples)},
  toptitle=0.5pt, bottomtitle=0.5pt,
  coltitle=black, colbacktitle=outtitle]
\small\sffamily
You are an expert at identifying and annotating hallucinations in visual question answering responses.

\textbf{Task:} The model gave an INCORRECT answer. Identify and annotate the hallucinated parts.

\textbf{Hallucination Types:} \textit{[Same 12 types as Stage 2a]}

\textbf{Dataset Category:} \texttt{\{dataset\_category\}}\\
\textbf{Focus on these types:} \texttt{\{primary\_types\}}

\textbf{Input:}\\
-- Question: \texttt{\{question\}}\\
-- Ground Truth Answer: \texttt{\{ground\_truth\}}\\
-- Model Answer (Incorrect): \texttt{\{model\_answer\}}

\textbf{Requirements:}\\
1.~Identify parts that differ from the ground truth.\\
2.~Use format: \texttt{<hallucination type="TYPE">hallucinated content</hallucination>}.\\
3.~Tags must not be nested or overlapping. Adjacent tags with the SAME type MUST be merged. Adjacent tags with DIFFERENT types are allowed.\\
4.~Each tag should cover a natural semantic unit --- single word, short phrase (2--3 words), or medium phrase (4--5 words). Do not over-split.\\
5.~Maintain text consistency after tag insertion.\\
6.~Diversify hallucination types; do not use a single type throughout.\\
7.~Vary tag lengths; mix single-word and short-phrase tags.\\
8.~Only merge adjacent hallucinated words when they belong to the same type.

\textbf{Output:} The model answer with hallucination annotations.
\end{tcolorbox}

\begin{tcolorbox}[breakable,colback=judgefill, colframe=black!30, boxrule=0.3pt,
  sharp corners, left=0.025\textwidth, right=0.025\textwidth, top=1pt, bottom=1pt,
  fonttitle=\footnotesize\bfseries,
  title={Stage 3: Quality Assessment},
  toptitle=0.5pt, bottomtitle=0.5pt,
  coltitle=black, colbacktitle=judgetitle]
\small\sffamily
You are an expert linguistic quality evaluator. Your task is to evaluate the quality of a generated answer for a Visual Question Answering (VQA) task.

\textbf{Important:} The answer contains hallucination tags in the format \texttt{<hallucination type="TYPE">content</hallucination>}. When evaluating quality, you should \textbf{IGNORE the actual content inside the hallucination tags} (as it represents intentionally injected errors). Focus only on evaluating the overall text structure, fluency, and coherence.

\textbf{Evaluate on three dimensions:}\\
1.~\textbf{Clarity} (1--5): Is the answer easy to understand? Are the sentences well-structured?\\
2.~\textbf{Fluency} (1--5): Is the language natural and grammatically correct? Does the text flow smoothly?\\
3.~\textbf{Coherence} (1--5): Does the answer structure make sense? Is the text internally consistent in terms of formatting and flow?

\textbf{Input:}\\
-- Question: \texttt{\{question\}}\\
-- Answer: \texttt{\{answer\}}

\textbf{Instructions:}\\
1.~Analyze the answer based on the three dimensions above.\\
2.~Ignore the factual correctness of content inside hallucination tags --- those are intentionally wrong.\\
3.~Evaluate the linguistic quality, structure, and readability of the overall text.\\
4.~Assign a score from 1 to 5 for each dimension.\\
5.~Provide a brief explanation for your scores.\\
6.~Determine if the sample is PASS or FAIL. PASS criteria: all scores $\geq$ 4.

\textbf{Output:} \texttt{<analysis>}...\texttt{</analysis>}, \texttt{<scores>} Clarity/Fluency/Coherence: [1--5] \texttt{</scores>}, \texttt{<decision>} PASS or FAIL with reason \texttt{</decision>}.
\end{tcolorbox}

\begin{tcolorbox}[breakable,colback=feedfill, colframe=black!30, boxrule=0.3pt,
  sharp corners, left=0.025\textwidth, right=0.025\textwidth, top=1pt, bottom=1pt,
  fonttitle=\footnotesize\bfseries,
  title={Stage 4: Consistency Verification},
  toptitle=0.5pt, bottomtitle=0.5pt,
  coltitle=black, colbacktitle=feedtitle]
\small\sffamily
You are an expert data quality auditor for a Hallucination Dataset. Your task is to verify the accuracy and correctness of hallucination tags injected into an answer.

The answer contains tags in the format: \texttt{<hallucination type="TYPE">content</hallucination>}.

\textbf{Allowed Hallucination Types:}\\
\textit{Perception-based:} Object, OCR, Numerical\_Attribute, Color\_Attribute, Shape\_Attribute, Spatial\_Attribute.\\
\textit{Reasoning-based:} Logical\_Error, Calculation\_Error, Knowledge\_Error, Query\_Misunderstanding, Numerical\_Relation, Spatial\_Relation.

\textbf{Verify four aspects:}\\
1.~\textbf{Syntax:} Are the tags properly formatted? (Opening and closing tags match, type attribute exists.)\\
2.~\textbf{Type Correctness:} Is the type attribute one of the 12 allowed types listed above?\\
3.~\textbf{Content Logic:} Does the content inside the tag actually represent a hallucination (error) compared to the Original Answer as ground truth?\\
4.~\textbf{Context:} Does the surrounding text flow naturally?

\textbf{Input:}\\
-- Question: \texttt{\{question\}}\\
-- Original Answer (Ground Truth): \texttt{\{original\_answer\}}\\
-- Hallucinated Answer (Target): \texttt{\{hallucinated\_answer\}}

\textbf{Instructions:}\\
1.~Compare the Hallucinated Answer against the Original Answer.\\
2.~Check every hallucination tag in the target answer.\\
3.~Verify if the tagged content contradicts the original answer or introduces an error consistent with its type.\\
4.~Verify if the non-tagged parts are consistent with the original answer (unless they were modified to support the hallucination).

\textbf{Output Format:}\\
\texttt{<analysis>} Analysis of the tags and their logical correctness compared to the original answer. \texttt{</analysis>}\\
\texttt{<validation>} Syntax\_Valid: [Yes/No], Types\_Valid: [Yes/No], Logic\_Valid: [Yes/No] \texttt{</validation>}\\
\texttt{<decision>} Result: [PASS/FAIL]. Reason: [Brief reason. If FAIL, specify which tag or issue caused it.] \texttt{</decision>}
\end{tcolorbox}

\subsection{Evaluation Prompt}
\label{app:prompt-diagnosis}

HalluScope-4B/8B use the following three-part prompt for hallucination diagnosis at inference time. The system prompt defines the 12 hallucination types with their descriptions and specifies a strict output format requiring tagged text wrapped in \texttt{<Tagged\_Text>} delimiters. The user prompt supplies the image, question, and model response to be analyzed.

\begin{tcolorbox}[breakable,colback=sysfill, colframe=black!20, boxrule=0.3pt,
  sharp corners, left=0.025\textwidth, right=0.025\textwidth, top=1pt, bottom=1pt,
  fonttitle=\footnotesize\bfseries, title=System Prompt,
  toptitle=0.5pt, bottomtitle=0.5pt,
  coltitle=black, colbacktitle=systitle]
\small\sffamily
You are a precise \textbf{hallucination detector and classifier} for multimodal large language models. Your task is to analyze images and model responses to \textbf{identify hallucinations}. You must \textbf{tag each hallucination with its specific type}.

\textbf{Hallucination Types} (use exactly these type names in tags):

\textbf{Perception-based} Hallucinations (can be directly verified from the image):\\
-- Object: Incorrect identification of objects in visual content.\\
-- OCR: Failure in text recognition processes within images.\\
-- Numerical\_Attribute: Misinterpretation of numerical values in visual elements.\\
-- Color\_Attribute: Errors in identifying the color.\\
-- Shape\_Attribute: Misrecognition of object shapes.\\
-- Spatial\_Attribute: Errors in recognizing the position, orientation, or distance of the object.

\textbf{Reasoning-based} Hallucinations (require inference beyond direct perception):\\
-- Logical\_Error: Errors in reasoning, such as incorrect causal relationships or conflicts in inference steps.\\
-- Calculation\_Error: Errors in mathematical operations (e.g., addition, subtraction, equation solving).\\
-- Knowledge\_Error: Applies incorrect domain knowledge or makes unrealistic inferences (e.g., violating common sense or physical laws).\\
-- Query\_Misunderstanding: Provides incorrect or irrelevant answers due to misunderstanding the query.\\
-- Numerical\_Relation: Misinterpreting the numerical relationship between objects (e.g., misreading proportions or quantities).\\
-- Spatial\_Relation: Misunderstanding the spatial, orientation, or distance relationships between objects.

\textbf{IMPORTANT OUTPUT FORMAT REQUIREMENTS:}\\
1.~Start with EXACTLY this line: ``Here is the response with hallucinated content tagged:''\\
2.~Then use \texttt{<Tagged\_Text>} tags to wrap the tagged response\\
3.~Inside \texttt{<Tagged\_Text>} tags:\\
-- Output the original response with \textbf{ONLY} \texttt{<hallucination type="TYPE">} tags added\\
-- \textbf{DO NOT modify or change any words} in the original response\\
-- ONLY add \texttt{<hallucination type="TYPE">} tags around hallucinated content\\
-- Each tag must include the \textbf{type attribute} specifying the hallucination category\\
-- If there are \textbf{no hallucinations}, output the original text exactly as is\\
4.~End with \texttt{</Tagged\_Text>}\\
5.~DO NOT add any other text, analysis, or explanation\\
6.~ANY OTHER FORMAT WILL BE REJECTED

Example Input:\\
prompt given to the model: describe the image\\
model's response: The bright red sports car is parked near a lake.

Correct Output Format:\\
Here is the response with hallucinated content tagged:\\
\texttt{<Tagged\_Text>}\\
The \texttt{<hallucination type="Color\_Attribute">}bright red\texttt{</hallucination>} \texttt{<hallucination type="Object">}sports\texttt{</hallucination>} car is \texttt{<hallucination type="Spatial\_Attribute">}parked near a lake\texttt{</hallucination>}.\\
\texttt{</Tagged\_Text>}

INCORRECT Outputs (DO NOT DO THESE):\\
-- Any text before ``Here is the response with hallucinated content tagged:''\\
-- Any text between the header and \texttt{<Tagged\_Text>}\\
-- Any text after \texttt{</Tagged\_Text>}\\
-- Any explanatory text or analysis\\
-- Any modification to the original text\\
-- Any additional formatting or tags besides \texttt{<hallucination type="TYPE">}

Remember:\\
-- Tag each hallucination \textbf{separately and specifically} with its type\\
-- Keep the content of the original text \textbf{strictly unchanged}\\
-- Use \textbf{exactly the type names} listed above (with underscores)\\
-- \textbf{Perception-based} types (Object, OCR, Numerical\_Attribute, Color\_Attribute, Shape\_Attribute, Spatial\_Attribute): Use when the error can be directly verified from the image\\
-- \textbf{Reasoning-based} types (Numerical\_Relation, Spatial\_Relation, Logical\_Error, Calculation\_Error, Knowledge\_Error, Query\_Misunderstanding): Use when the error involves comparison, reasoning, or inference beyond direct observation
\end{tcolorbox}

\begin{tcolorbox}[breakable,colback=userfill, colframe=black!20, boxrule=0.3pt,
  sharp corners, left=0.025\textwidth, right=0.025\textwidth, top=1pt, bottom=1pt,
  fonttitle=\footnotesize\bfseries, title=User Prompt,
  toptitle=0.5pt, bottomtitle=0.5pt,
  coltitle=black, colbacktitle=usertitle]
\small\sffamily
\texttt{<image>}\\
Here is the prompt given to the model:\\
\texttt{\{question\}}

Here is the model's response:\\
\texttt{\{answer\}}

Please analyze the image and add \texttt{<hallucination>} tags to any hallucinated content in the model's response. Remember to tag each hallucinated content separately!
\end{tcolorbox}

\subsection{Feedback Prompt Templates}
\label{app:prompt-feedback}

All four feedback modes share the same system prompt and differ only in the user message. The diagnostic granularity increases progressively: Self-Refine simply informs the target model that its previous answer was incorrect without any external diagnosis; CRITIC-Det provides detection-level feedback by marking hallucinated spans with structured correction instructions but without type classification; CRITIC-Cls further incorporates type annotations into both span marking and correction instructions; and HalluScope uses HalluScope-8B to generate interpretable explanations that inform the target model what is wrong, why it is wrong, and how to correct it.

\begin{tcolorbox}[breakable,colback=sysfill, colframe=black!20, boxrule=0.3pt,
  sharp corners, left=0.025\textwidth, right=0.025\textwidth, top=1pt, bottom=1pt,
  fonttitle=\footnotesize\bfseries, title=System Prompt,
  toptitle=0.5pt, bottomtitle=0.5pt,
  coltitle=black, colbacktitle=systitle]
\small\sffamily
You are a careful visual question answering assistant. You will be given your previous answer with hallucination annotations marked. Please re-examine the image, consider the marked hallucinations, and provide a corrected answer.
\end{tcolorbox}

\begin{tcolorbox}[breakable,colback=feedfill, colframe=black!20, boxrule=0.3pt,
  sharp corners, left=0.025\textwidth, right=0.025\textwidth, top=1pt, bottom=1pt,
  fonttitle=\footnotesize\bfseries,
  title={Self-Refine},
  toptitle=0.5pt, bottomtitle=0.5pt,
  coltitle=black, colbacktitle=feedtitle]
\small\sffamily
\textbf{User:} Your previous answer to this question was incorrect. Please try again.

Please:\\
1.~Carefully re-examine the image\\
2.~Think step by step about what you actually see\\
3.~Provide a corrected answer

Question: \texttt{\{question\}}

Now please provide your corrected answer based on what you actually observe in the image.
\end{tcolorbox}

\begin{tcolorbox}[breakable,colback=judgefill, colframe=black!20, boxrule=0.3pt,
  sharp corners, left=0.025\textwidth, right=0.025\textwidth, top=1pt, bottom=1pt,
  fonttitle=\footnotesize\bfseries,
  title={CRITIC-Det},
  toptitle=0.5pt, bottomtitle=0.5pt,
  coltitle=black, colbacktitle=judgetitle]
\small\sffamily
\textbf{User:} Your previous answer to this question had some issues. A reviewer has identified the following problems: 

\textit{[Generated by \texttt{generate\_structured\_summary\_from\_detection()}:]}

Question: \texttt{\{question\}}. Please re-examine the image carefully, correct the issues listed above, and provide a new answer.
\end{tcolorbox}

\begin{tcolorbox}[breakable,colback=userfill, colframe=black!20, boxrule=0.3pt,
  sharp corners, left=0.025\textwidth, right=0.025\textwidth, top=1pt, bottom=1pt,
  fonttitle=\footnotesize\bfseries,
  title={CRITIC-Cls},
  toptitle=0.5pt, bottomtitle=0.5pt,
  coltitle=black, colbacktitle=usertitle]
\small\sffamily
\textbf{User:} Your previous answer to this question had some issues. A reviewer has identified the following problems: 

\textit{[Generated by \texttt{generate\_structured\_summary()}:]}

Question: \texttt{\{question\}}. Please re-examine the image carefully, correct the issues listed above, and provide a new answer.
\end{tcolorbox}

\begin{tcolorbox}[breakable,colback=outfill, colframe=black!20, boxrule=0.3pt,
  sharp corners, left=0.025\textwidth, right=0.025\textwidth, top=1pt, bottom=1pt,
  fonttitle=\footnotesize\bfseries,
  title={HalluScope},
  toptitle=0.5pt, bottomtitle=0.5pt,
  coltitle=black, colbacktitle=outtitle]
\small\sffamily
\textbf{User:} Your previous answer to this question had some issues. Here is feedback from a reviewer:

\texttt{\{suggestion\}}

\textit{[The \texttt{suggestion} is generated by HalluScope-8B using the \texttt{<Suggestion>} variant of the evaluation prompt: 1--3 sentences explaining what is wrong, why it is wrong based on the image, and how to correct it.]}

Question: \texttt{\{question\}}. Please re-examine the image carefully based on the feedback above and provide a corrected answer.
\end{tcolorbox}

\end{document}